\documentclass[nohyperref]{article}

\usepackage{microtype}
\usepackage{graphicx}
\usepackage{booktabs} 

\usepackage{hyperref}
\usepackage{wrapfig}
\usepackage{caption}
\usepackage{subcaption}
\usepackage{multirow}

\newcommand{\NAME}{\texttt{SMERF}}

\usepackage[accepted]{icml2022}

\usepackage{amsmath}
\usepackage{amsfonts}
\usepackage{amssymb}
\usepackage{mathtools}
\usepackage{amsthm}

\usepackage[capitalize,noabbrev]{cleveref}

\theoremstyle{plain}

\theoremstyle{definition}

\theoremstyle{remark}

\usepackage[textsize=tiny]{todonotes}

\icmltitlerunning{Sanity Simulations for Saliency Methods}

\begin{document}

\twocolumn[
\icmltitle{Sanity Simulations for Saliency Methods}




\begin{icmlauthorlist}
\icmlauthor{Joon Sik Kim}{cmu}
\icmlauthor{Gregory Plumb}{cmu}
\icmlauthor{Ameet Talwalkar}{cmu}
\end{icmlauthorlist}

\icmlaffiliation{cmu}{Machine Learning Department, Carnegie Mellon University, Pittsburgh, USA}
\icmlcorrespondingauthor{Joon Sik Kim}{joonkim@cmu.edu}

\icmlkeywords{saliency methods, interpretability}

\vskip 0.3in
]



\printAffiliationsAndNotice{}  

\begin{abstract}
Saliency methods are a popular class of feature attribution explanation methods that aim to capture a model's  predictive reasoning by identifying ``important'' pixels in an input image.
However, the development and adoption of these methods are hindered by the lack of access to ground-truth model reasoning, which prevents accurate evaluation.
In this work, we design a synthetic benchmarking framework, \NAME{}, that allows us to perform ground-truth-based evaluation while controlling the complexity of the model's reasoning. 
Experimentally, \NAME{} reveals significant limitations in existing saliency methods and, as a result, represents a useful tool for the development of new saliency methods.
\end{abstract}

\section{Introduction}

\begin{figure}[t]
    \centering
    \includegraphics[width=\columnwidth]{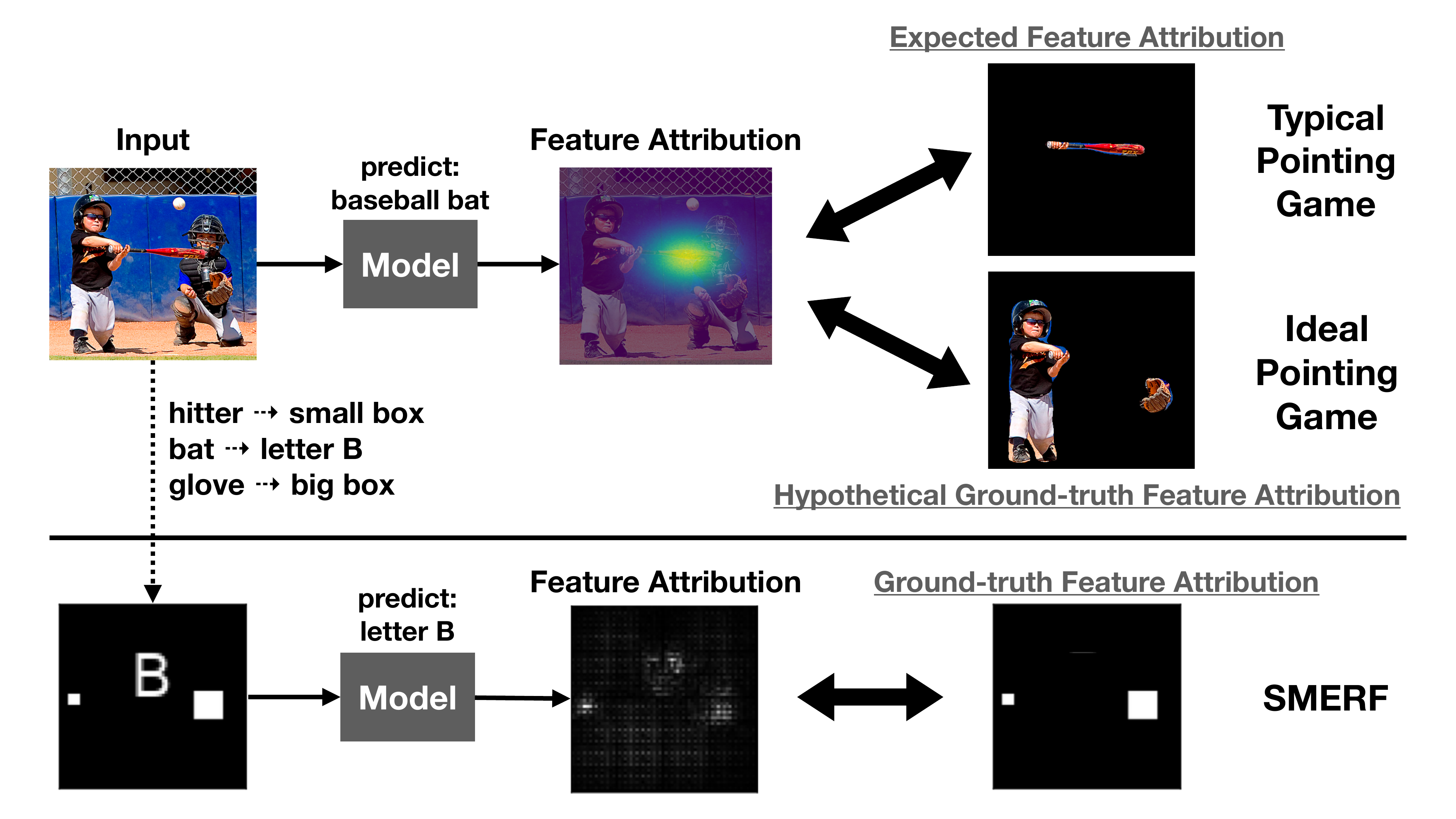}
    \caption{\textbf{Top.} Existing pointing game evaluations do not have access to ground-truth feature attributions but instead rely on \textit{expected} feature attributions. 
    For instance, while a model trained to identify a baseball bat is expected to rely on the baseball bat region of the image (top), it may rely on a more complex reasoning by using the presence of a hitter and a glove to identify a bat (bottom).
    \textbf{Bottom.} \NAME{} constructs a synthetic set of tasks that are stylized versions of real object classification tasks.
    Consider the task of identifying the letter `B' in an image, where the letter and the two boxes correspond to the bat, the hitter, and the glove, respectively.
    \NAME{} controls the model's underlying reasoning via simulations over different data distributions, providing ground-truth feature attributions used to evaluate saliency methods (in this example, a model is relying on two boxes to identify the letter).}
    \label{fig:pipeline}
\end{figure}

Saliency methods are a popular tool to help understand the behavior of machine learning models.
Given a model and an input image, these methods output a feature attribution that indicates which pixels they deem to be most ``important'' to the model's prediction~\cite{simonyan2013deep, sundararajan2017axiomatic, lundberg2017unified}.
Then, a natural question to ask is: \emph{how do we define ``important'' and subsequently evaluate the efficacy of these methods?}  

One intuitive approach to answering this question is to measure how well a saliency method locates the expected pixels of interest in the input image. 
In fact, this so-called “pointing game” ~\cite{zhang2018top} is one of the predominant evaluations used today~\cite{Zhou_2016_CVPR, selvaraju2017grad, chattopadhay2018grad, woo2018cbam, gao2019res2net, arun2020assessing}. 
Current versions of this evaluation rely on external knowledge to define an \emph{expected} feature attribution that highlights the region that a human would expect to be important for the given task.
Then, the quality of a saliency method is measured using the overlap between its output and this expected feature attribution by metrics such as Intersection-Over-Union (IOU) (See Section~\ref{sec:experiments}).

Unfortunately, this approach has two key limitations. 
First, the results are unreliable when the model’s ground-truth reasoning does not match human expectations, e.g., when the model is relying on spurious correlations. 
This is particularly problematic because detecting such discrepancies is one of the motivating use cases of saliency methods.   
Second, existing versions are based on relatively simple object classification tasks where we expect only a single region of the image, i.e., the object itself, to be relevant to the prediction. 
In practice, there exist more complex tasks, e.g., in medical imaging or autonomous driving, where considering interactions among multiple regions of the image may be necessary for the model to achieve high predictive accuracy. 

These two limitations highlight the same fundamental concern: we do not know a priori what or how complex the model’s reasoning will be, irrespective of how simple we think the underlying task is.  
For instance, the top panel of Figure~\ref{fig:pipeline} considers the seemingly simple task of identifying a baseball bat in an image. 
Based on the description of the task, we might expect the model to use \emph{simple reasoning}, which we define as relying on a single region of the image, e.g., the bat itself, to make its prediction. 
If this is the case, the expected feature attribution should highlight the bat only.
However, if the model actually uses more \emph{complex reasoning}, which we define as relying on interactions among multiple regions of the image, e.g., using the presence of a hitter and a glove to identify a bat, the actual ground-truth feature attribution should highlight the hitter and the glove, not the bat.
As illustrated through this example, the correct evaluation of saliency methods fundamentally depends on the model's ground-truth reasoning.

Consequently, we aim to address these key limitations by controlling the model's ground-truth reasoning. 
To do this, we start by generating synthetic images composed of simplified objects and consistent backgrounds that are stylized versions of real-world scenarios (e.g., Figure~\ref{fig:pipeline} Bottom).  
Then, by controlling the distribution and label of these images, we can induce and then verify a specific ground-truth reasoning for the model.  
By repeating this process for different levels of reasoning complexity, we build a benchmark called \textbf{S}imulated \textbf{M}od\textbf{E}l \textbf{R}easoning Evaluation \textbf{F}ramework (\NAME{}) that can evaluate saliency methods against ground-truth model reasoning.

\begin{figure}
    \centering
    \includegraphics[width=\columnwidth]{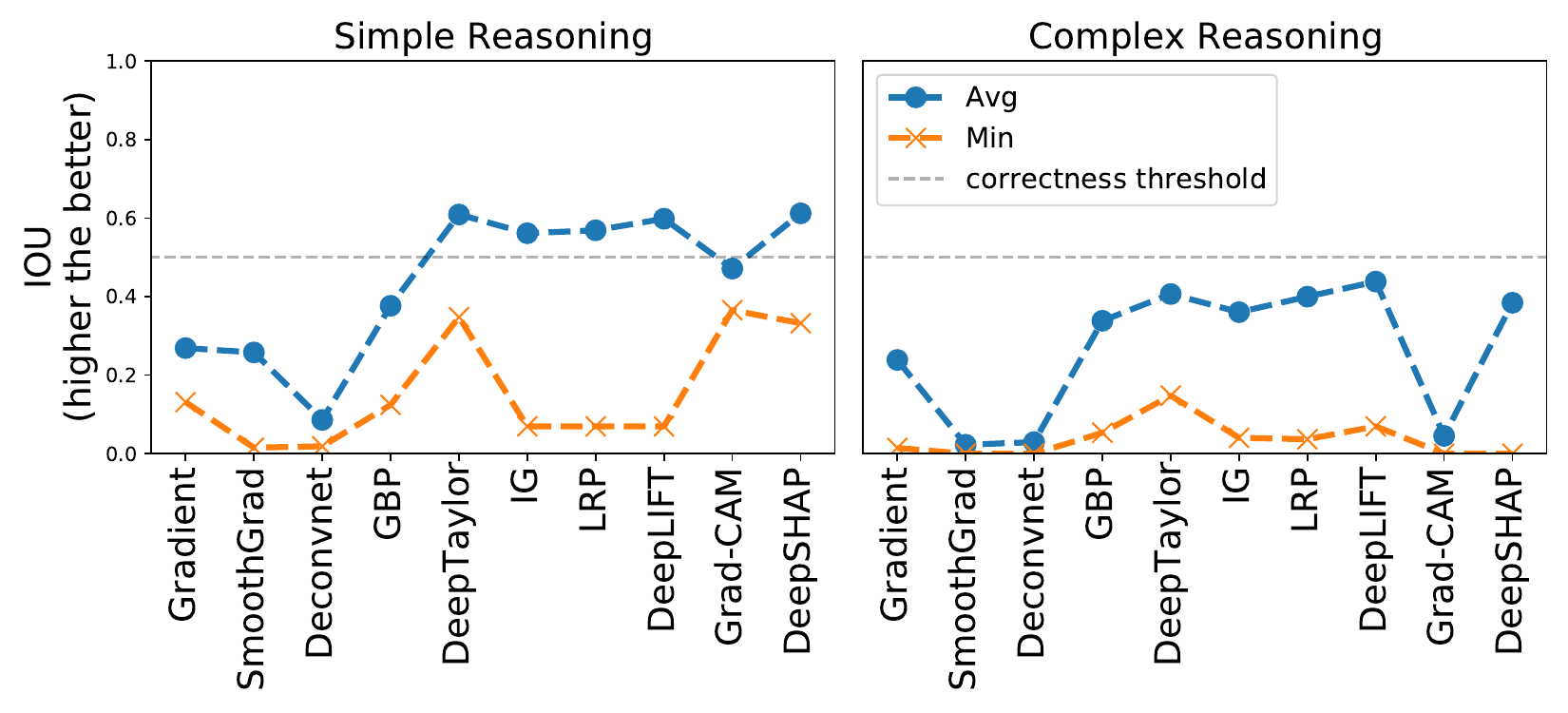}
    \caption{Summary of ground-truth-based evaluation of saliency methods via \NAME{}. 
    \textbf{Left.} In simple reasoning settings, where the model relies on a single region of the image to make its prediction, average performance (blue) is reasonably good for most of the methods. However, all methods still demonstrate failure cases as shown by minimum performance (orange) over various tasks. 
    \textbf{Right.} In more complex reasoning settings, where the model relies on interactions among multiple regions of the image, average performance drops with more acute failure cases.}
    \label{fig:general_results}
\end{figure}

Using \NAME{}, we consider seven distinct model reasoning settings with varying complexity, and perform an extensive evaluation of 10 leading saliency methods for each setting. 
Our analyses are summarized in Figure~\ref{fig:general_results} and discussed at  length throughout Section~\ref{sec:experiments}. We observe that for simple reasoning settings, leading saliency methods perform reasonably well on average, though still exhibit certain failure cases (Section~\ref{sec:simple_result1}).
We further observe clear performance degradation as we increase model reasoning complexity. Indeed, in all complex reasoning settings, none of the methods meet our (lenient) definition of correctness\footnote{While we view the IOU $> 0.5$ as a lenient definition of correctness in synthetic settings, this value is commonly used in practice when evaluating on real tasks~\cite{everingham2015pascal, wang2019pseudo}.}, and all of them demonstrate acute failure cases (Section~\ref{sec:complex_result1}, \ref{sec:model-choice}).

Our results highlight major limitations of existing saliency methods, especially given the relative simplicity of \NAME{}’s synthetic evaluation tasks.  
We further illustrate how \NAME{}'s synthetic evaluations translate to more natural images, by presenting qualitatively similar yet generally worse results on analogous reasoning tasks that leverage natural image backgrounds instead of synthetic ones (Section~\ref{sec:real-background}).

\section{Related Work}

\textbf{Pointing Game Evaluation.} 
The pointing game, which measures how well a saliency method identifies the relevant regions of an image, is one of the predominant ways to evaluate the efficacy of these methods~\cite{Zhou_2016_CVPR, selvaraju2017grad, chattopadhay2018grad, woo2018cbam, gao2019res2net, arun2020assessing}.  
Many existing pointing game evaluations lack access to ground-truth model reasoning but instead rely on expected feature attributions generated by domain experts.
Intuitively, this might appear to be reasonable by observing that the model has high test accuracy and concluding that it must be using the correct reasoning.
However, datasets often contain spurious correlations and, as a result, a model may be able to achieve high test accuracy using incorrect reasoning.  
Consequently, these evaluations have confounded the correctness of the explanation with the correctness of the model.  
\NAME{} eliminates this confounding factor by leveraging the model's ground-truth reasoning, which allows us to demonstrate that several methods previously deemed to be effective are in fact sometimes ineffective for more complex model reasoning. 

\citet{yang2019benchmarking, adebayo2020debugging} try to address this same limitation using semi-synthetic datasets where the ground-truth reasoning is known by combining the object from one image with the background from another.  
Both analyses are based on the simple reasoning setting and, in that setting, our results roughly corroborate theirs.   
However, our analysis extends to more complex reasoning settings and demonstrates that methods that worked in the simple reasoning setting mostly perform much worse in these settings. 
It is important to consider the complex reasoning setting because we do not know how complex the model's reasoning is in practice (e.g., a model may rely on a spurious correlation and use complex reasoning for a simple task). 

A concurrent work by \citet{zhou2022feature} 
introduces a similar semi-synthetic pipeline for testing saliency methods, where a family of image manipulations is applied to the input so that the ground-truth impact of specific features on the model prediction is known.
Whereas \citet{zhou2022feature} focuses on model reasoning that relies on a single artificial feature in the image, we establish a complementary criteria for saliency methods by focusing on a more diverse set of model reasoning complexity induced by interactions among different features in the image.

\textbf{Direct Criticisms of Saliency Methods.} 
\citet{adebayo2018sanity} uses two sanity checks that measure the statistical relationship between a saliency method and the model's parameters or the data it was trained on.  
They found that only a few methods (i.e., Gradient \cite{simonyan2013deep} and Grad-CAM \cite{selvaraju2017grad}) passed these tests. 
\citet{kindermans2019reliability} similarly tests several saliency methods for input invariance and finds that the Gradient satisfies the property while other methods generally do not. 
While \NAME{} is orthogonal to such types of analyses, it demonstrates that even methods that pass these tests have failure cases (as shown in Figure~\ref{fig:general_results}).  
\citet{shah2021input} suggests that a model's adversarial robustness impacts how well the Gradient is able to correctly focus only on the relevant features. 
\NAME{} verifies this observation for simple reasoning, and further shows that the problem persists in complex reasoning settings even for robust models for all tested methods.

\textbf{Other Evaluations.}
Beyond the pointing game, several proxy metrics have been proposed to evaluate saliency methods~\cite{bach2015pixel, ancona2018towards, alvarez2018, hooker2019benchmark}.
Additionally, \citet{liu2021synthetic} introduces a benchmarking framework based on synthetic datasets sampled from different types of Gaussian distributions to evaluate the methods on these proxy metrics. 
However, \citet{Tomsett_2020} shows that several popular proxy metrics inherently depend on subtle hyperparameters that are not well understood and that this leads to analyses with inconsistent results.
The unreliability of these popular proxy metrics further emphasizes the advantage of using a more intuitive evaluation like \NAME{}.
Similar setups with humans in the loop are also being introduced to measure the user's perception of the feature attributions in detecting model biases~\cite{sixt2021users}.

\section{Methods}

\NAME{} is a synthetic evaluation benchmark where several types of ground-truth model reasoning, ranging from simple to complex, are generated to test a saliency method's ability to recover them. 
We first describe several types of model reasoning that are motivated by real-world examples and then captured in \NAME{}'s synthetic family of datasets called \texttt{TextBox}.
We then explain the data generation and training process in \NAME{} with \texttt{TextBox}.
Additional details are in Appendix \ref{sec:appdx_training}.

\subsection{Types of Simple and Complex Model Reasoning}
\label{sec:types_of_reasoning}

We are interested in evaluating the performance of saliency methods in capturing both simple and complex model reasoning, where the complexity of reasoning is defined based on whether the model relies on a \textit{single} or \textit{multiple} regions of the image. 
We next describe three different ways in which a model may exhibit simple or complex reasoning characterized by the model's reliance on the true set of features ($X$) and/or the set of spurious features ($F$).

\begin{figure}[t]
    \centering
    \includegraphics[width=0.85\columnwidth]{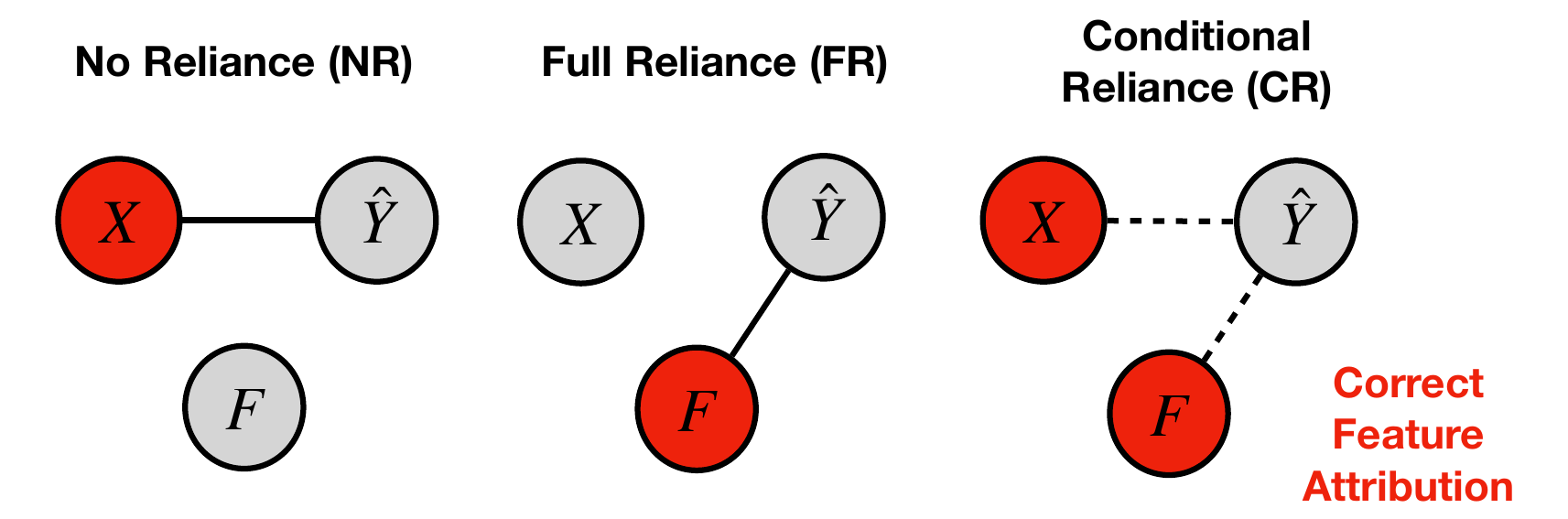}
    \caption{\small{Given true features $X$ and spurious features $F$ in the training data, the model may exhibit simple or complex reasoning depending on how it relies on $X$ and/or $F$. No-Reliance (NR) and Full-Reliance (FR) denote settings where the model relies solely on $X$ or $F$, respectively. Conditional Reliance (CR) denotes settings where the model depends on both $X$ and $F$. \NAME{} allows us to control the model's reasoning and to thus evaluate saliency methods against the ground-truth feature attribution (denoted in red) derived from this underlying reasoning.}}
    \label{fig:types}
\end{figure}

Consider the model trained to detect a baseball bat from Figure~\ref{fig:pipeline}. 
The model may correctly rely on the true set of features (e.g. the bat), without relying on spurious features (e.g. the glove or the hitter).  
We denote models that exhibit no reliance on spurious features as the 
\textit{No-Reliance (NR)} setting (Figure~\ref{fig:types}, left).
Another model could instead depend on the existence of the glove and the hitter, thus fully relying on spurious features, a setting we call \textit{Full-Reliance (FR)} (Figure~\ref{fig:types}, middle). 
Finally, the model may rely on both the true and the spurious sets of features (Figure~\ref{fig:types} right), a setting we denote as \textit{Conditional-Reliance (CR)}.
For instance, a model may learn to rely on the glove only when the hitter is present, but otherwise on the bat itself for the prediction.

CR by default prescribes complex reasoning due to its conditional nature. 
In contrast, the complexity of model reasoning for NR and FR depends on how many objects are included in $X$ and $F$: simple when $X$ and $F$ each consists of features corresponding to a single object, and complex when they consist of multiple objects. 
It is notable that existing pointing game evaluations in the literature are performed with respect to a simple reasoning under the assumption that the model exhibits NR, with $X$ being the single object of interest~\cite{Zhou_2016_CVPR, selvaraju2017grad}. 
Moreover, previous controlled setups for evaluating saliency methods were limited to simple model reasoning in the FR setting, e.g., setting $F$ as the background and $X$ as a single object of interest~\cite{yang2019benchmarking, adebayo2020debugging}.

 \begin{figure}[t]
    \centering
    \includegraphics[width=0.35\columnwidth]{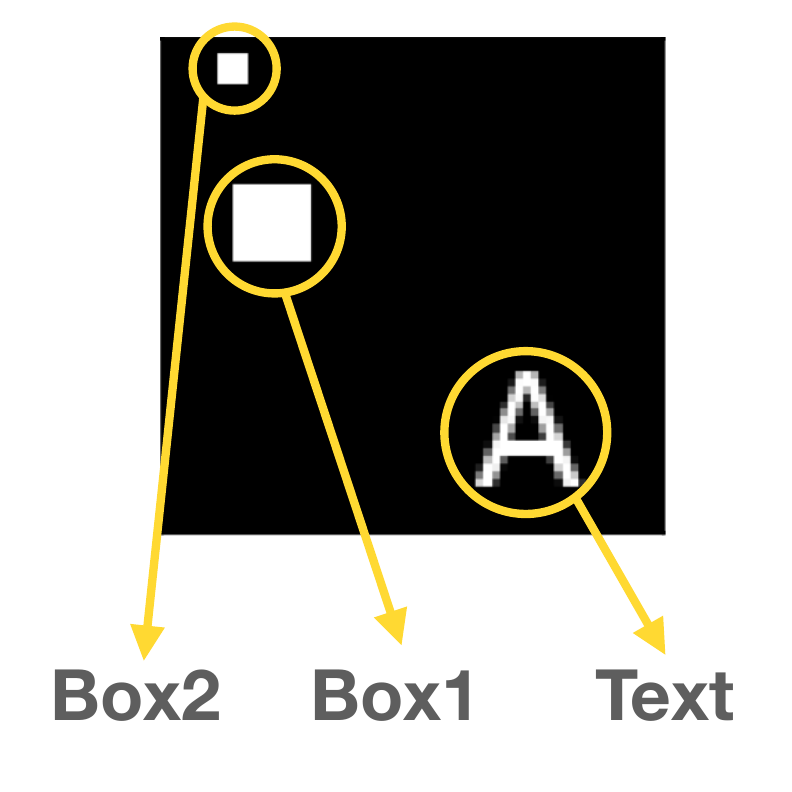}
    \caption{\small{Features in the \texttt{TextBox} datasets.}}
    \label{fig:data_example}
\end{figure} 

\NAME{} instantiates simple and complex model reasoning across these three settings by creating a family of datasets called \texttt{TextBox}. 
These datasets all consist of 64-by-64 pixel images with black background that include three types of white objects in random locations of each image (Figure~\ref{fig:data_example}): \textbf{Text}, ‘A’, ‘B’; \textbf{Box1}, a 10-by-10 box; and \textbf{Box2}, a 4-by-4 box. 
\NAME{} simulates the objects' relationship with the labels (as in Figure~\ref{fig:types}) to control the model reasoning for $X=\textrm{Text}$, $F=\{\textrm{Box1, Box2}\}$.
Note that the choice of these features are arbitrary -- they can be replaced by other shapes, colors, and backgrounds.


\subsection{Training Models with Ground-truth Reasoning}
\label{sec:generating}

\begin{figure*}[t]
    \centering
    \includegraphics[width=0.9\textwidth]{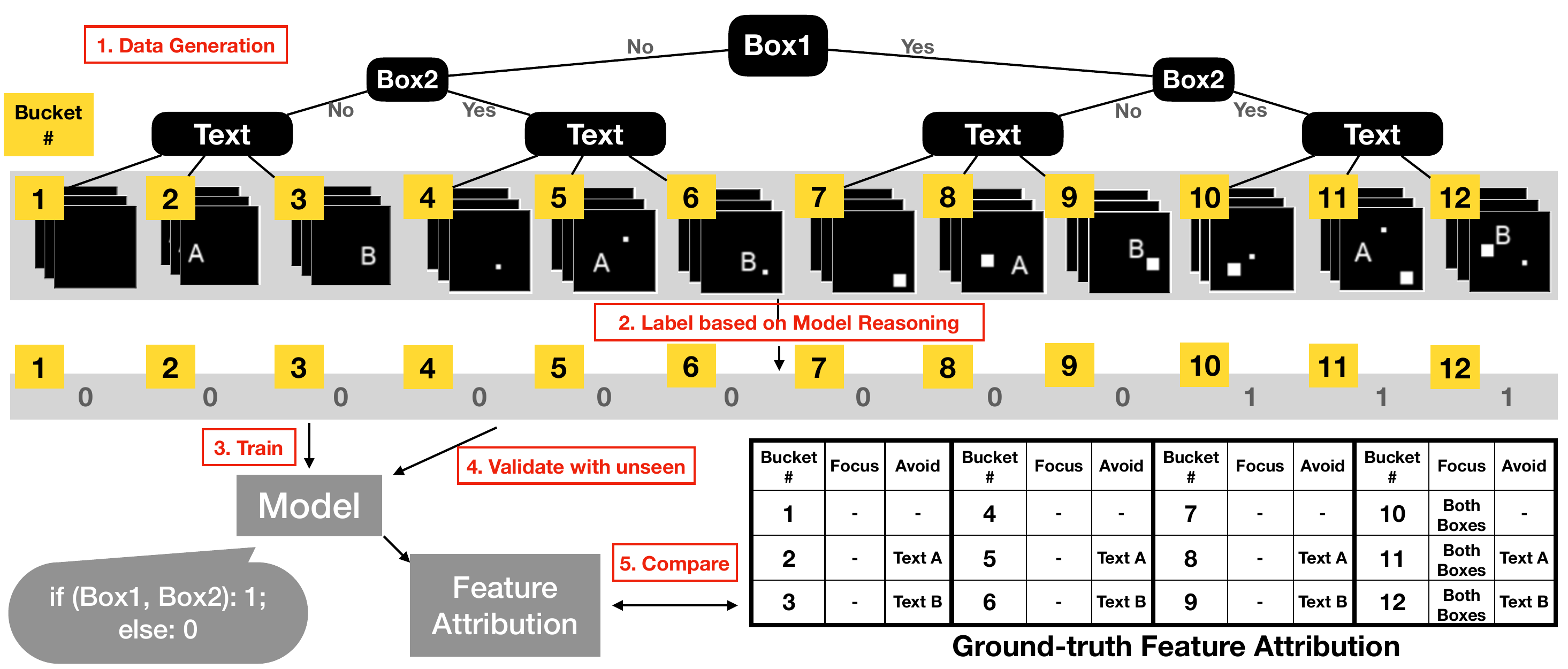}
    \caption{\small{Workflow of \NAME{} for a model with FR on Box1 and Box2, where the model predicts 1 if both boxes are present, otherwise 0 (Complex-FR in Table~\ref{tab:dataset_info}). Twelve buckets of images each composed of different sets of features are generated. These are then labeled according to the model reasoning.
    The model is then trained/validated on samples from each bucket. 
    The ground-truth feature attribution should focus/avoid certain objects in the image, as labels depend on specific objects only, e.g. labels do not depend on Text, but only on both Boxes (shown in the table). Feature attributions from saliency methods are compared against this ground-truth.}}
    \label{fig:smerf_fr}
\end{figure*}

\begin{table*}[h!]
    \centering
    \resizebox{\textwidth}{!}{\begin{tabular}{c c c c c c c c}
        Name & \textbf{Simple-FR} & \textbf{Simple-NR} & \textbf{Complex-FR} & \textbf{Complex-CR1} & \textbf{Complex-CR2} & \textbf{Complex-CR3} & \textbf{Complex-CR4} \\ \hline \hline
        \begin{tabular}{@{}c@{}} Reasoning \\ (How buckets  \\ are labeled) \end{tabular} & \begin{tabular}{@{}c@{}} if B1: 1; \\ otherwise: 0 \end{tabular} & \begin{tabular}{@{}c@{}} if T=A: 0; \\ if T=B: 1 \end{tabular} & \begin{tabular}{@{}c@{}} if B1 \& B2: 1; \\ otherwise: 0 \end{tabular} & \begin{tabular}{@{}c@{}} if B2 \& B1: 1; \\ if B2 \& NO B1: 0; \\ if NO B2 \& T=A: 0; \\ if NO B2 \& T=B: 1 \end{tabular}& \begin{tabular}{@{}c@{}} if B1 \& B2: 1; \\ if B1 \& NO B2: 0; \\ if NO B1 \& T=A: 0; \\ if NO B1 \& T=B: 1 \end{tabular}& \begin{tabular}{@{}c@{}} if B2 \& T=A: 0; \\ if B2 \& T=B: 1; \\ if NO B2 \& B1: 1; \\ if NO B2 \& NO B1: 0 \end{tabular} & \begin{tabular}{@{}c@{}} if B1 \& T=A: 0; \\ if B1 \& T=B: 1; \\ if NO B1 \& B2: 1; \\ if NO B1 \& NO B2: 0 \end{tabular} \\ \hline
        \# of Buckets & 12 & 8 & 12 & 10 & 10 & 10 & 10 \\ \hline
        \begin{tabular}{@{}c@{}} ID of Buckets \\ Labeled 0\end{tabular} & 1,2,3,4,5,6 & 2,5,8,11 & 1,2,3,4,5,6,7,8,9 & 2,4,5,6,8 & 2,5,7,8,9 & 1,2,3,5,11 & 1,2,3,8,11 \\ \hline
         \begin{tabular}{@{}c@{}} ID of Buckets \\ Labeled 1\end{tabular}& 7,8,9,10,11,12 & 3,6,9,12 & 10,11,12 & 3,9,10,11,12 & 3,6,10,11,12 & 6,7,8,9,12 & 4,5,6,9,12\\ \hline
        \begin{tabular}{@{}c@{}} ID of Undefined \\ Buckets \end{tabular} & None & 1,4,7,10 & None & 1,7 & 1,4 & 4,10 & 7,10\\ \hline
        \begin{tabular}{@{}c@{}} Objects \\ to Focus \end{tabular} & \begin{tabular}{@{}c@{}} 1,2,3,4,5,6: N \\ 7,8,9,10,\\11,12: B1 \end{tabular} & \begin{tabular}{@{}c@{}} 2,3,5,6,8,\\ 9,11,12: T \end{tabular} & \begin{tabular}{@{}c@{}} 1,2,3,4,5\\,6,7,8,9: N \\ 10,11,12: B1,B2\end{tabular} & \begin{tabular}{@{}c@{}} 2,3,8,9: T \\ 4,5,6: B2 \\ 10,11,12: B1, B2\end{tabular} & \begin{tabular}{@{}c@{}} 2,3,5,6: T \\ 7,8,9: B1 \\ 10,11,12: B1, B2\end{tabular} & \begin{tabular}{@{}c@{}} 1,2,3: N \\ 5,6,11,12: B2,T \\ 7,8,9:B1  \end{tabular} & \begin{tabular}{@{}c@{}} 1,2,3: N \\ 4,5,6: B2 \\ 8,9,11,12:B1,T  \end{tabular}\\ \hline
        \begin{tabular}{@{}c@{}} Objects \\ to Avoid \end{tabular}  & \begin{tabular}{@{}c@{}} 1: N \\ 2,3,8,9: T \\ 4,10: B2 \\ 5,6,11,12: B2, T \end{tabular} & \begin{tabular}{@{}c@{}} 2,3: N \\ 5,6: B2 \\ 8,9: B1 \\ 11,12: B1,B2 \end{tabular} & \begin{tabular}{@{}c@{}} 1,4,7,10: N \\ 2,3,5,6,11,12: T \\  \end{tabular} & \begin{tabular}{@{}c@{}} 2,3,4,10: N \\ 8,9: B1 \\  5,6,11,12: T \end{tabular} & \begin{tabular}{@{}c@{}} 2,3,7,10: N \\ 5,6: B2 \\ 8,9,11,12: T \end{tabular} & \begin{tabular}{@{}c@{}} 1,5,6,7: N \\ 2,3,8,9: T \\ 11,12: B1 \end{tabular} & \begin{tabular}{@{}c@{}} 1,4,8,9: N \\ 2,3,5,6: T \\ 11,12: B2 \end{tabular} \\ \hline
    \end{tabular}}
    \caption{\small{The seven model reasoning settings considered in the experiments (Section~\ref{sec:experiments}). Each column represents a model reasoning setting that belongs to one of the three categories depicted in Figure~\ref{fig:types}. \textbf{B1} stands for Box1, \textbf{B2} stands for Box2, \textbf{T} stands for Text, and \textbf{N} stands for None (see Figure~\ref{fig:data_example}). 
    Depending on the reasoning, there are different numbers of buckets that belong to the positive/negative classes, and different objects that feature attributions should focus on/avoid. Bucket ID numbers are taken from Figure~\ref{fig:smerf_fr}, which corresponds to the setting described in the third column of this table (Complex-FR). See Appendix~\ref{sec:appdx_dataset} for more details.
    }}
    \label{tab:dataset_info}
\end{table*} 

\NAME{} first creates an appropriate training dataset for a particular desired model reasoning. 
Specifically, \NAME{} starts by generating 12 \textit{buckets}\footnote{The total number of buckets depends on the cardinality of $X$ and $F$, as we create buckets for all possible $(X,F)$ value pairs; hence for \texttt{TextBox} datasets we consider 12 buckets: there are three different values for Text (Nothing, `A', or `B'), two for Box1, and two for Box2, resulting in a total of 12 distinct combinations (Appendix~\ref{sec:appdx_dataset}).} of images (as shown in Figure~\ref{fig:smerf_fr}), where each bucket contains images with particular $X$ and $F$, and an associated label designated by the desired model reasoning.
More formally, considering the joint distribution $p(X, F, Y)$, each bucket will be composed of images with features $X=x$, $F=f$ along with the label $y \sim p(Y|X=x, F=f)$ determined by the specified model reasoning, with different images in each bucket varying by the location of the features. 
A convolutional neural network\footnote{We consider a shallow CNN, AlexNet~\cite{alexnet2012}, and VGG16~\cite{simonyan2014very} for the model architectures. The results in Section~\ref{sec:experiments} are from the shallow CNN, and we observe similar results from other deep architectures, as reported in Section~\ref{sec:model-choice} and Appendix~\ref{sec:appdx_other_architectures}.}
is trained on the entire set of buckets, which is then validated with unseen data points from each bucket to ensure that the ground-truth model reasoning has been properly learned.
Because the data distribution is simulated, we can generate arbitrary number of images from different buckets which differ only in terms of a single
feature and use them to confidently verify that specific features are responsible for the model’s prediction. 
Repeating this for all possible subsets of features effectively ensures that the model follows the intended reasoning.
Ground-truth feature attributions are derived from this verified model reasoning and are later used for evaluating saliency methods.  

Figure~\ref{fig:smerf_fr} depicts the steps of dataset generation and model training/validation for complex model reasoning with FR. 
In this example, we want the labels to depend only on the presence of both boxes, thus providing positive labels only for the three buckets that include both boxes (buckets 10-12), and negative labels for the nine buckets that include at most one box (buckets 1-9).
We verify that the model learns the desired model reasoning, as it achieves near-perfect accuracy on unseen samples from each bucket.
The ground-truth model reasoning provides ground-truth feature attribution shown in the table of Figure~\ref{fig:smerf_fr}, defining regions the saliency methods should focus and/or avoid for images from each buckets.
This information will later be used to evaluate the feature attributions obtained from the saliency methods.

\section{Experiments}
\label{sec:experiments}

We use \NAME{} and the \texttt{TextBox} datasets to show the performance of leading saliency methods for different types of model reasoning presented in Table~\ref{tab:dataset_info} with varying complexity (the example from Figure~\ref{fig:smerf_fr} corresponds to Complex-FR in the third column).
For simple reasoning (Section~\ref{sec:simple_result1}), we find that saliency methods perform reasonably well (with a few exceptions), which is generally consistent with previous pointing game evaluations.
However, we observe a general trend of decreasing performance when input images become more ``saturated'' (i.e., filled with more objects), even when the model's underlying reasoning does not change. 
For complex reasoning (Section~\ref{sec:complex_result1}), the average performance for all methods decreases, with several failure cases due to methods focusing more on irrelevant objects. 
As a result, feature attributions qualitatively become indistinguishable across different model reasoning, raising practical concerns since users in general do not know the type of reasoning being used (and in fact would potentially rely on saliency methods to get this information). 
We further show that a similar trend is present when varying factors like model architecture choice and robust training (Section~\ref{sec:model-choice}). And when the images contain more natural backgrounds (Section~\ref{sec:real-background}), even lower worst-case performance is observed.
Additional details about the results are in Appendix \ref{sec:appdx_exp_result}.

\textbf{Saliency Methods and Baselines.} 
We use a modified version of the open-source library \texttt{iNNvestigate}\cite{JMLR:v20:18-540} which includes several implementations of leading saliency methods. 
We use the following methods: Gradient (G)~\cite{simonyan2013deep}, SmoothGradients (SG)~\cite{smilkov2017smoothgrad}, DeConvNet (DCN)~\cite{zeiler2014visualizing}, Guided Backpropagation (GBP)~\cite{springenberg2015striving}, Deep Taylor Decomposition (DT)~\cite{montavon2017explaining}, Input*Gradient (I*G)~\cite{shrikumar2017learning}, Integrated Gradients (IG)~\cite{sundararajan2017axiomatic}, and Layerwise Relevance Propagation (LRP)~\cite{bach2015pixel} (four variations: LRP-z, LRP-$\epsilon$, LRP-A$\flat$, LRP-B$\flat$), DeepLIFT (DL)~\cite{shrikumar2017learning} (two variations: DL-RC (using Reveal-Cancel rule), DL-R (using Rescale rule)), Grad-CAM (G-CAM)~\cite{selvaraju2017grad}, and DeepSHAP (D-SHAP)~\cite{lundberg2017unified}. 
We also add some simple baselines, like Random (random-valued feature attribution) and Edge-detection (Edge), both of which are model-independent and therefore are not useful in understanding the model reasoning. 

\begin{figure*}[t]
    \centering
     \begin{subfigure}[b]{0.7\textwidth}
     \centering
          \includegraphics[width=\textwidth]{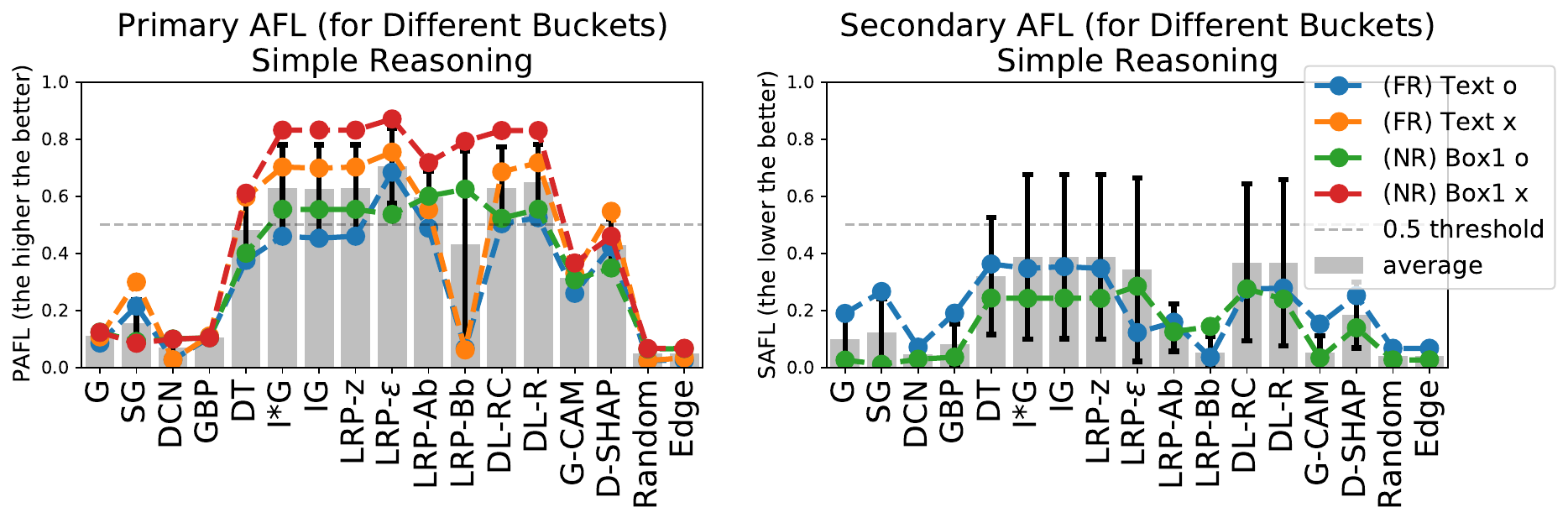}
         \caption{PAFL and SAFL for Simple Reasoning}
         \label{fig:simple_afls}
     \end{subfigure} 
     \hfill
     \begin{subfigure}[b]{0.29\textwidth}
         \centering
          \includegraphics[width=\columnwidth]{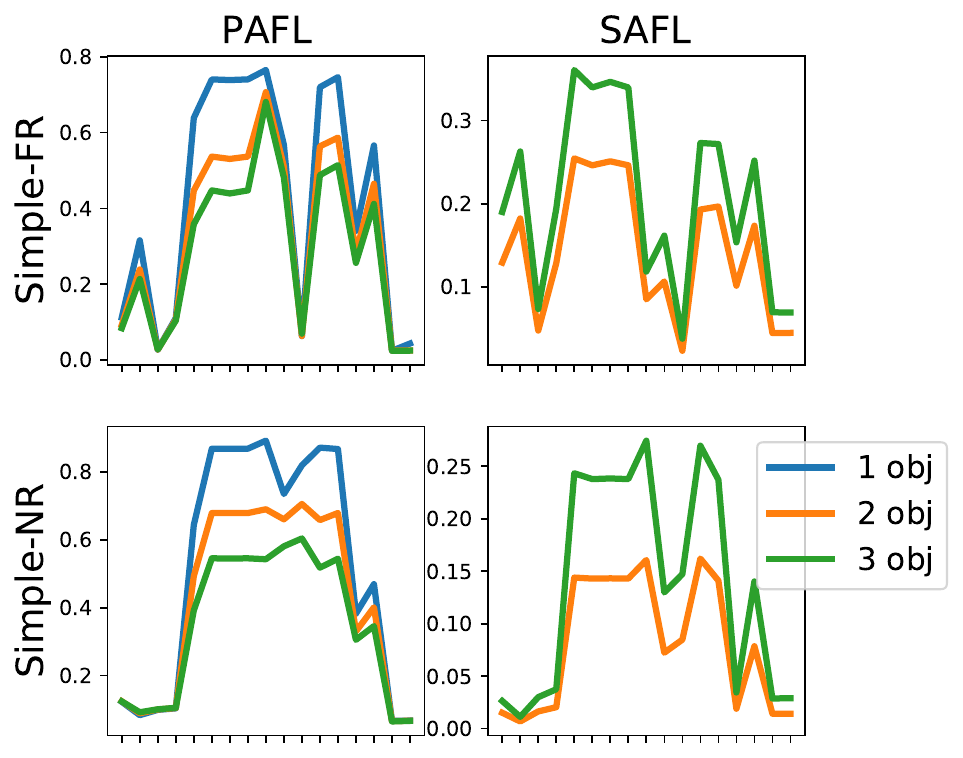}
         \caption{PAFL and SAFL based on the number of objects in the image (x-axis is the same as Figure~\ref{fig:simple_afls})}
         \label{fig:simple_afls_obj}
     \end{subfigure}
    \caption{\textbf{(a)} PAFL (left) and SAFL (right) for Simple-FR and Simple-NR (Table~\ref{tab:dataset_info}). 
    The black vertical lines indicate the standard deviation across different buckets. 
    Most methods perform well on average, with reasonable PAFL and low SAFL.
    The methods on the left part of the plot (G, SG, CGN, CBP) are the exception. 
    The colored lines indicate average performance for different buckets that have either Text or Box1 (i.e. the irrelevant features) present (o) or absent (x). 
    \textbf{(b)} PAFL decreases and SAFL increases as the number of objects (regardless of their relevance to the prediction) increases, which indicates that the methods perform worse as images become ``saturated.''
    }
    \label{fig:simple_results}
\end{figure*}

\textbf{Evaluation Metrics.}
Typical pointing game evaluations measure the performance of saliency methods with the Intersection-Over-Union (IOU) metric~\cite{everingham2015pascal, Zhou_2016_CVPR}.
This metric computes the ratio of the area of the intersection to the area of the union of the binary-masked feature attribution and the ground-truth feature attribution. 
The binary-masked feature attribution is obtained by first blurring the original feature attribution averaged across the three color channels, followed by thresholding the pixel intensity to select the top-$K$ pixels, where $K$ is equal to the number of pixels that contain the object of interest.
Given the popularity of this metric, we perform extensive experiments with it: Figure~\ref{fig:general_results} summarizes the results, which are consistent with those of our main results.

However, we find that IOU loses information about raw attribution values on extra objects when thresholding to generate the binary-masked feature attribution for evaluation.
It is therefore more likely to disregard non-trivial signals from extra objects in the image (see Appendix~\ref{sec:appdx_eval_metric_afl}).

To address these issues, we also consider a previously introduced~\cite{wang2020score} but unnamed metric that we call \emph{Attribution Focus Level (AFL)}.
This metric quantifies the proportion of the total attribution values that are assigned to the region of interest. 
Given the raw, normalized feature attribution values, it is the sum of values assigned to pixels inside the region of interest. 
Intuitively, values near 1 indicate a stronger level of focus on the region of interest.
We define a threshold value of 0.5, chosen to indicate that  more than half of the total attribution values is focused on the object, to roughly distinguish good and bad performance in terms of AFL.

To better account for the relationship among multiple features in the image, we use two types of AFL: 
(1) \textit{Primary AFL (PAFL)}, which measures the level of focus on the \textit{correct} (primary) features that are relevant for the prediction (corresponding to objects to focus in Table~\ref{tab:number_training} and Figure~\ref{fig:smerf_fr}), and (2) \textit{Secondary AFL (SAFL)}, which measures the same quantity for the \textit{incorrect} (secondary) features that are irrelevant for the prediction (corresponding to objects to avoid in Table~\ref{tab:number_training} and Figure~\ref{fig:smerf_fr}, excluding background). 
Notably, the sum of PAFL and SAFL is upper bounded by 1, so PAFL $>$ 0.5 implies that PAFL $>$ SAFL, which further indicates that the feature attribution correctly focuses more on the relevant features than the irrelevant ones (our definition of ``success''). 
Conversely, when SAFL $>$ PAFL, the feature attribution incorrectly focuses more on irrelevant regions and is thus considered undesirable (our definition of ``failure'').

\subsection{Simple Reasoning Setting}
\label{sec:simple_result1} 

\begin{figure}[t]
    \centering
    \includegraphics[width=0.8\columnwidth]{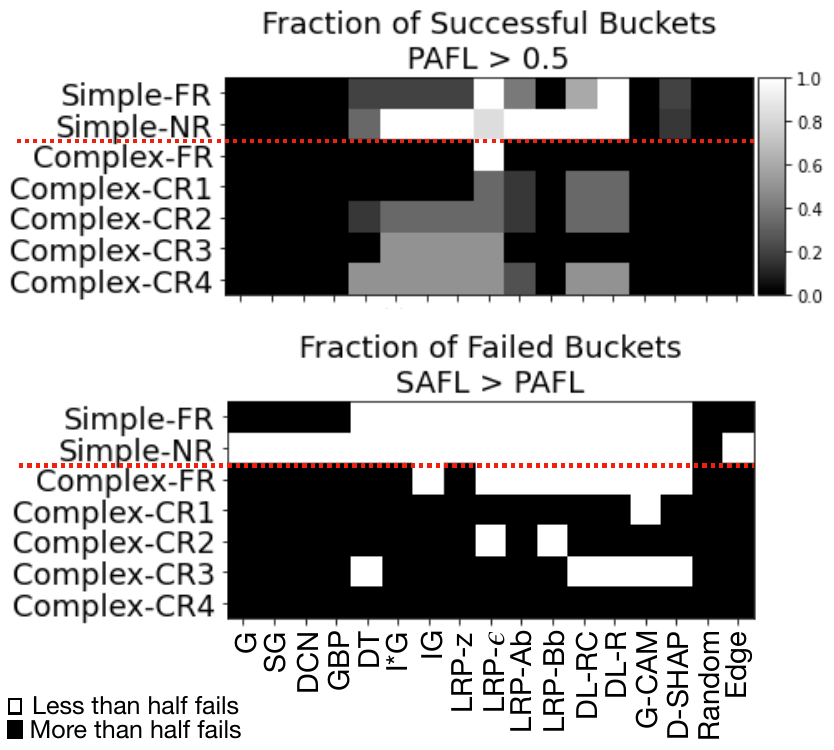}
    \caption{\small{\textbf{Top:} The fraction of buckets for which the methods are successful (i.e., PAFL $>$ 0.5).
    Most success cases (white) are concentrated on simple reasoning setting.
    \textbf{Bottom:} The cases where the method fails due to wrong focus (i.e., SAFL $>$ PAFL) on more than half of the buckets for each reasoning (colored with black). 
    Most failure cases are concentrated on complex reasoning, which aligns with increasing SAFL observed in Figure~\ref{fig:complex_afls}.}}
    \label{fig:caseview}
\end{figure}

\begin{figure*}[t]
    \begin{subfigure}[b]{0.70\textwidth}
         \centering
         \includegraphics[width=\textwidth]{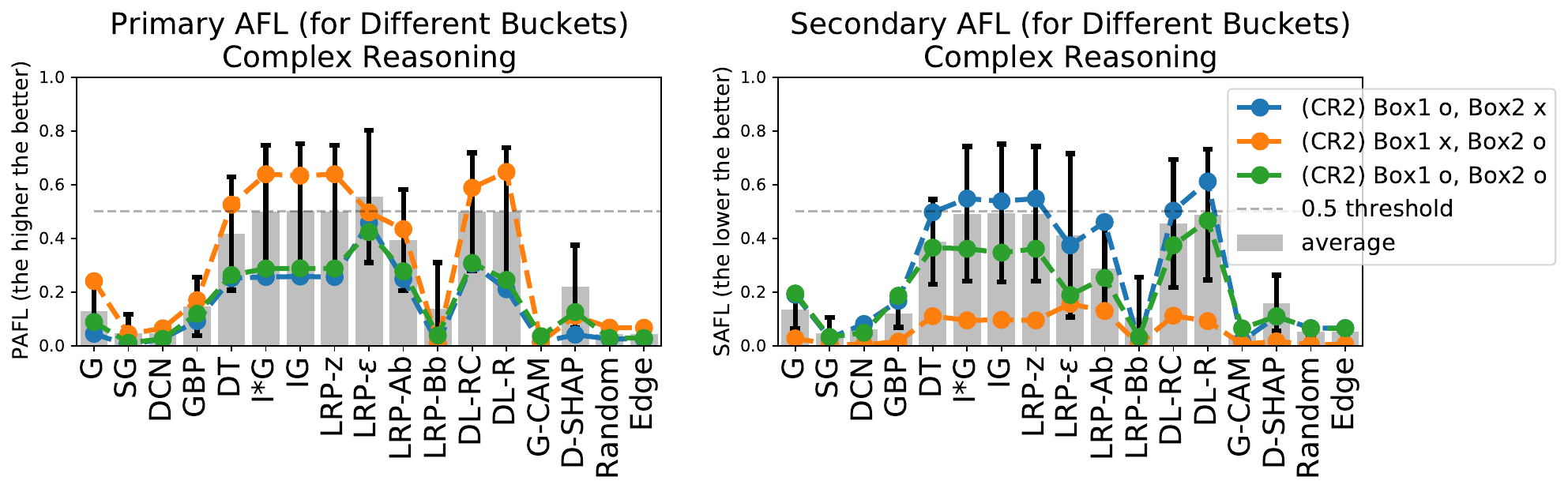}
         \caption{PAFL and SAFL for Complex Reasoning}
         \label{fig:complex_afls}
     \end{subfigure} 
    \hfill
    \begin{subfigure}[b]{0.29\textwidth}
         \centering
         \includegraphics[width=\textwidth]{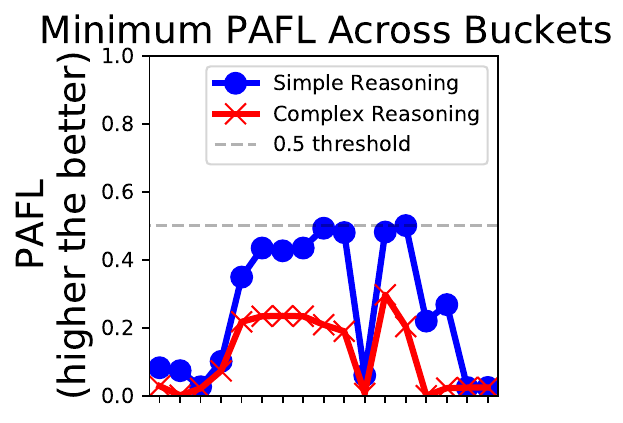}
         \caption{Minimum PAFL over buckets on simple (blue circle) vs. complex (red x) reasoning (x-axis is the same as Figure~\ref{fig:complex_afls}).}
         \label{fig:complex_afls_min}
     \end{subfigure} 
    \caption{\small{\textbf{(a)} PAFL and SAFL for complex reasoning. 
    Average performance on PAFL is mostly lower than 0.5, with worse performance compared to the results for simple reasoning. 
    Further, SAFL increases significantly (to the point where it is on par with PAFL for some methods). 
    Per-bucket performance variation for Complex-CR2 (from Table~\ref{tab:dataset_info}) is plotted with colored lines, each color corresponding to the methods' performance on a bucket with either Box1 or Box2 present (o) or absent (x). 
    \textbf{(b)} Worst-case buckets show worse PAFL values in complex reasoning (red) with a sharp drop from simple reasoning setting (blue). 
    }}
    \label{fig:complex_result}
\end{figure*} 

For a high-level understanding of how the methods generally perform, we plot PAFL and SAFL averaged across all buckets for simple reasoning instantiated with Simple-FR and Simple-NR (grey vertical bars in Figure~\ref{fig:simple_afls}). 
We observe that most of the methods, except for G, SG, DCN, and GBP, perform reasonably well, with PAFL exceeding the 0.5 correctness threshold and SAFL being lower.
This reasonable level of performance aligns with what existing evaluations in the literature have shown with simple reasoning~\cite{selvaraju2017grad, adebayo2020debugging}. 

Despite their reasonable performance in average, we observe a trend that PAFL decreases and SAFL increases as more objects are visible in the image, even though the model reasoning remains simple. 
The colored lines in Figure~\ref{fig:simple_afls} exemplify this trend by showing per-bucket performance for two different buckets in Simple-FR (blue and orange) and Simple-NR (green and red) each. 
For both reasoning, all methods show lower PAFL on buckets where an irrelevant object (Text for Simple-FR in blue and Box1 for Simple-NR in green) is present, compared to buckets where that object is absent (orange and red). 
This means that part of the feature attribution originally assigned to the relevant object shifts to the irrelevant ones when they are visible in the image.
Figure~\ref{fig:simple_afls_obj} verifies this trend for all buckets containing different number of visible objects: PAFL from buckets with fewer objects strictly upper bounds PAFL from buckets with more objects, while SAFL increases along with the number of objects. 
These variations in AFL based on the number of objects in the image lead to non-trivial variance across buckets as shown with black vertical error bars (standard deviation) in Figure \ref{fig:simple_afls}.  
Qualitative examples confirm this undesirable dependence of methods' AFL values on the number of objects in the image (Appendix~\ref{sec:appdx_qual_results}). 

To better view the success/failure cases, we record the fraction of buckets that contain both relevant and irrelevant features for which each method is considered successful and indicate it with a color ranging from black (0) to white (1) (Figure~\ref{fig:caseview} Top).
For simple reasoning (the first two rows), methods in the middle (I*G through DL) have relatively higher success rates overall.
Among these, DL-R is the only method that succeeds in \textit{all} buckets for \textit{both} types of simple reasoning. 
Methods like G, SG, DCN, and GBP fail to be successful for all buckets.

\subsection{Complex Reasoning Setting}
\label{sec:complex_result1}

We consider 5 types of complex reasoning in Table~\ref{tab:dataset_info}: Complex-FR, and Complex-CR1 through Complex-CR4.  
Compared to the simple reasoning case, PAFL drops for all methods (grey vertical bars in Figure~\ref{fig:complex_afls}). 
Methods with strong performance in simple reasoning settings (IG, LRP, and DL) narrowly meet the correctness threshold for complex reasoning settings, while those with decent performance (G-CAM and D-SHAP) suffer from bigger drops in PAFL.
Methods like G, SG, DCN, and GBP, which showed weak performance in simple reasoning settings, continue to show low PAFL. 

In addition to lower average PAFL overall, we see an increase in SAFL for complex reasoning. 
In some cases, SAFL approaches the 0.5 threshold, demonstrating a clear failure by focusing on the wrong object(s) more (grey bars in Figure~\ref{fig:complex_afls} right). 
This is immediately verified in Figure~\ref{fig:caseview} (top), where with a single exception (LRP-$\epsilon$ on Complex-FR), \textit{none} of the methods are successful in all of the buckets for complex reasoning (rows below the red dotted line).
Figure~\ref{fig:caseview} (bottom) further demonstrates 
that the majority of the buckets show failures for complex reasoning (colored with black), contrary to simple reasoning (mostly white).

We next measure the worst-case performance of different methods (i.e., by evaluating the worst-performing buckets for each method, as we also visualized in Figure~\ref{fig:general_results}). 
We observe that the worst-case PAFL for complex reasoning is much lower than that for simple reasoning (Figure~\ref{fig:complex_afls_min}).
Qualitative samples from these worst-performing buckets speak for the low PAFL with clear lack of focus on the relevant features (Appendix~\ref{sec:appdx_qual_results}).

Moreover, we observe that per-bucket performance variation is more extreme in complex reasoning settings.  
For instance, per-bucket performance on Complex-CR2 (blue, orange, and green lines in Figure~\ref{fig:complex_afls}) shows that most methods are successful in a bucket where only Box2 is present (orange), while for other buckets with only Box1 (blue) or both Boxes (green), all methods clearly fail. 
Such variation is also visible across other types of complex reasoning (Appendix~\ref{sec:appdx_complex_reasoning_perbucket}).
These altogether contribute to higher variance of performance for complex reasoning settings (indicated with black vertical error bars in Figure~\ref{fig:complex_afls}), higher than the simple reasoning settings (Figure~\ref{fig:simple_afls}). 

\begin{figure}[t]
    \centering
    \includegraphics[width=\columnwidth]{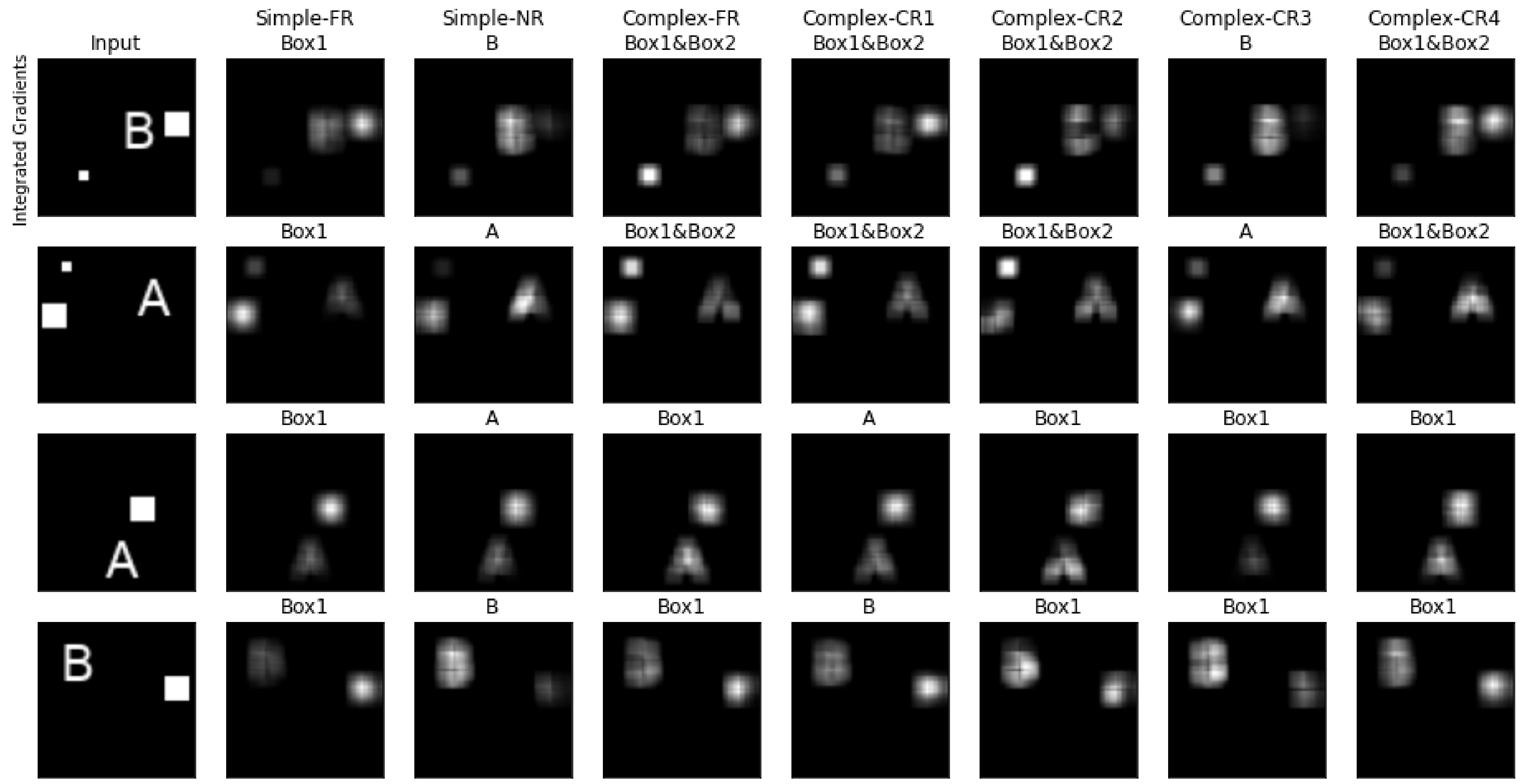}
    \caption{\small{Feature attributions from Integrated Gradients for different model reasoning on four inputs from different buckets, labeled with their relevant features to highlight. Essentially all objects in the image are highlighted, making it difficult to identify which type of model reasoning is used for each column.}}
    \label{fig:identify}
\end{figure}

Finally, we contextualize the aforementioned failure cases of saliency methods from a practitioners' perspective, who are not aware of the type of model reasoning used, but are relying on the feature attributions for this information.
By failing to point to only the correct set of features, these methods are likely to mislead practitioners. 
Figure~\ref{fig:identify} visualizes the feature attributions from IG (one of the best methods in our evaluation) for each model reasoning (see Appendix~\ref{sec:appdx_identify} for other methods).  
Essentially all objects are highlighted in all samples, and it is thus not clear how to discern the underlying model reasoning from any of them. 
For example, Box1 appears to be the most important object according to the feature attributions in the third row, even when this object is clearly not part of the model reasoning for Simple-NR (3rd column) and Complex-CR1 (5th column).

\subsection{Impact of Model Architecture and Robustness}
\label{sec:model-choice}

We next vary the model's architecture and its robustness and show that the trends we observed in previous sections persist. 

\begin{figure}[t]
    \centering
    \includegraphics[width=\columnwidth]{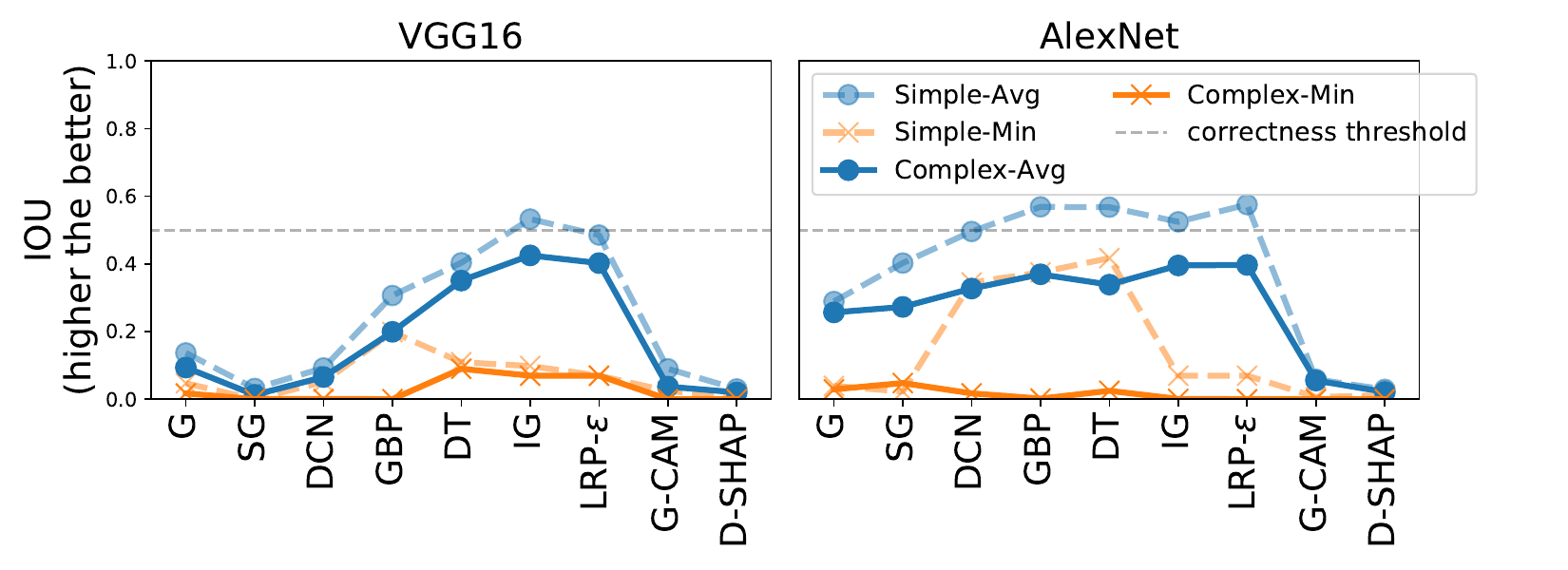}
    \caption{IOU results on VGG16 (Left) and AlexNet (Right). For both models, all methods show drops in both average (blue) and worst-case (orange) performance on complex reasoning (solid lines) compared to simple reasoning (dotted lines).}
    \label{fig:architecture-comparison}
\end{figure}

\noindent \textbf{Model Architecture.} We repeat the same set of experiments with AlexNet~\cite{alexnet2012} and VGG16~\cite{simonyan2014very} to confirm that the trends we observe for the saliency methods are not the artifact of model architecture choice. Figure~\ref{fig:architecture-comparison} shows similar trends we have observed so far: both the average (blue lines) and the worst-case performance drops (orange lines) as the reasoning becomes more complex (comparing solid against dotted lines).
For more details, see Appendix~\ref{sec:appdx_other_architectures}. 

\begin{figure}[t]
    \centering
    \includegraphics[width=\columnwidth]{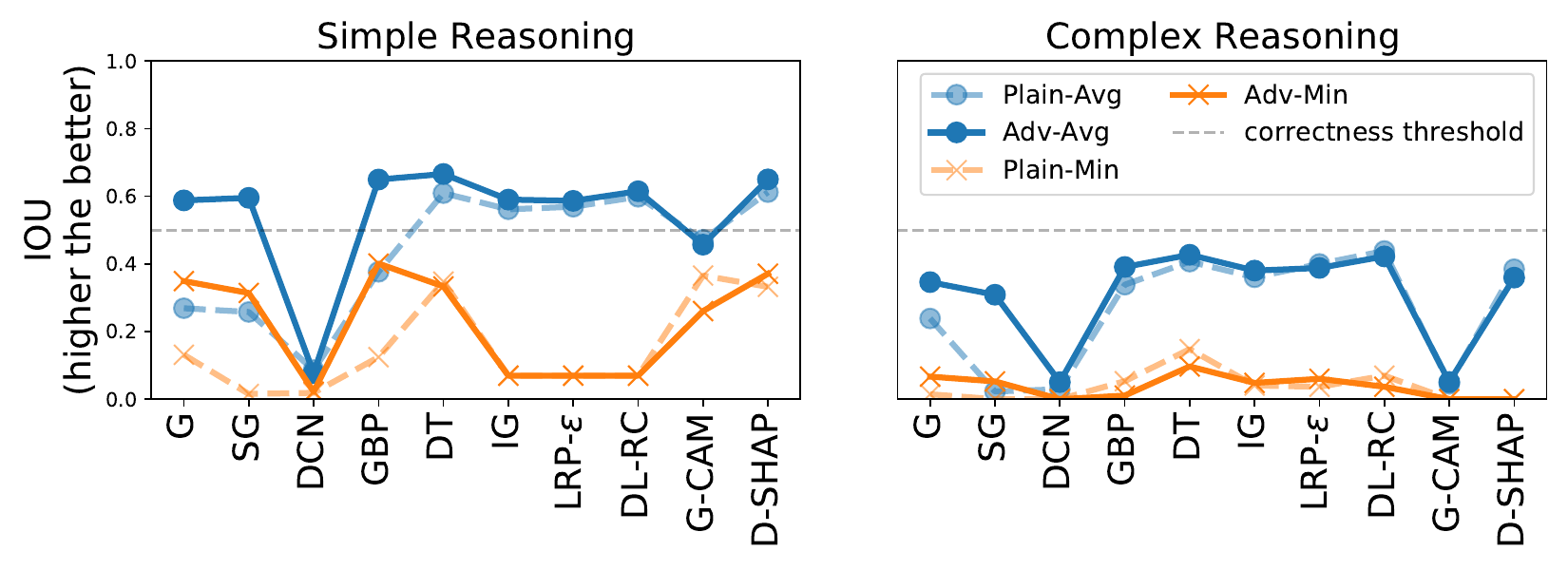}
    \caption{IOU results on robust models trained against PGD attacks (solid lines) and plain models (dotted lines, taken from Figure~\ref{fig:general_results}) for simple (left) and complex reasoning settings (right). Robust models show higher average performance for Gradient and SmoothGradient in simple reasoning compared to plain models, as suggested in \cite{shah2021input}. Nevertheless, there still are performance drops for complex reasoning across all methods.}
    \label{fig:adversarial_tests}
\end{figure}

\noindent \textbf{Adversarial Robustness.} It has been previously suggested that Gradient feature attribution applied on adversarially robust models tend to better ignore the signals from spurious objects in the image~\cite{shah2021input}. 
We verify that this trend for Gradient feature attribution is somewhat true, yet the general problem persists for most of the methods even for robust models~\cite{madry2018towards}, in both simple and complex reasoning settings (Figure~\ref{fig:adversarial_tests}). 
While Gradient (G) and SmoothGradient (SG)'s performance on robust models (solid lines) is higher compared to plain models (dotted lines) for simple reasoning, we still observe performance drops in complex reasoning throughout all methods. 
For more details, see Appendix~\ref{sec:appdx_adversarial}.

\subsection{Extending to Natural Image Backgrounds}
\label{sec:real-background}

\begin{figure}[t]
    \centering
    \includegraphics[width=0.6\columnwidth]{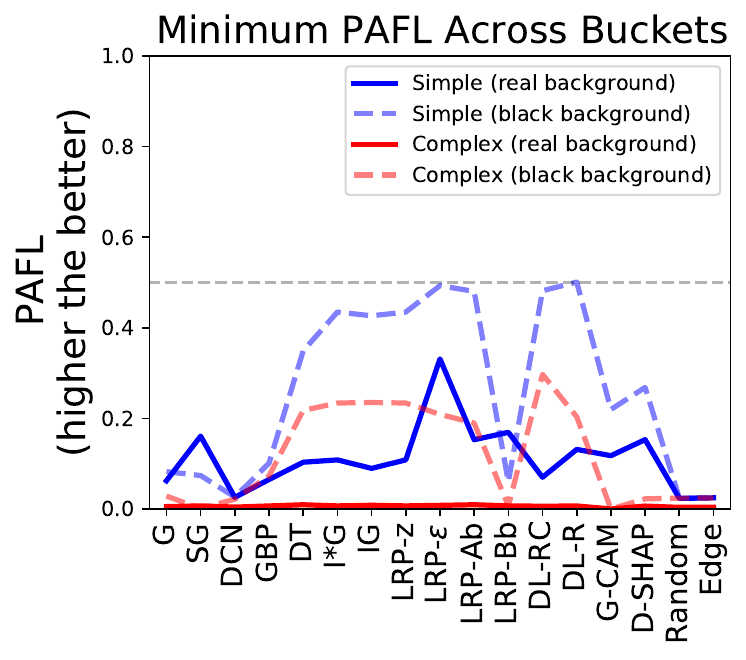}
    \caption{\small{Minimum PAFL comparison between cases with uniform black background (dotted lines) and real background (solid lines). We observe a decrease in performance when moving from simple (blue) to complex reasoning (red) even for the real background case. Due to more noise from the background, the overall PAFL values are lower for the real background case.}}
    \label{fig:comparison-backgrounds}
\end{figure}

In previous sections we focused on images with uniform black background.
With this black background, we aimed to make it easy for the model to identify the objects present in the image. 
We also expected that this simplistic background would improve the performances of the saliency methods\footnote{We also run experiments using random noise pixels for the background and observe performance drops (Appendix~\ref{sec:appdx_random_bg}).}.

However, we can simulate more realistic scenarios by replacing the black background with natural images.
To this end, we replace the background of the \texttt{TextBox} datasets with real images of baseball stadiums, chosen to simulate tasks that are more similar in spirit to the one depicted in the top panel of Figure~\ref{fig:pipeline}, sampled from the Places dataset~\cite{zhou2017places}, while still reasoning over the same objects (Text, Box1, Box2), and repeat the experiments in Sections~\ref{sec:simple_result1} and \ref{sec:complex_result1}.

Figure~\ref{fig:comparison-backgrounds} shows the summary of results on images with real backgrounds (solid lines), comparing them to what was observed in Sections~\ref{sec:simple_result1} and \ref{sec:complex_result1} on images with black backgrounds (dotted lines, identical to Figure~\ref{fig:complex_afls_min}). 
There are two types of performance drop observed. 
First, we observe a similar drop as we move from simple (blue solid line) to complex (red solid line) reasoning for both the black and real background settings. 
Second, we see a general performance drop in the real background setting as compared to the black background setting.
For instance, in the real background setting, all methods are far from the threshold even for simple reasoning settings (solid blue line), which differs from results in the black background setting (dotted blue line). See Appendix~\ref{sec:appdx_changing_bg} for more details.

Collectively, these results suggest that the performance of these saliency methods is likely to further deteriorate under even more realistic, noisier scenarios. They thus highlight the importance of consistent success in controlled (synthetic) settings as a stepping stone to success in real-world settings.

\section{Conclusion}
\label{sec:conclusion}

Using \NAME{}, we created seven stylized prediction tasks which revealed significant shortcomings of existing saliency methods, especially in their ability to recover complex model reasoning. 
We further corroborated the main results with additional results using natural image backgrounds, demonstrating similar qualitative trends but degraded quantitative performance. 
\NAME{} serves as a natural benchmark to evaluate saliency methods' correctness, and our results suggest that it can roughly provide an optimistic upper bound of a method's performance in more complicated real-world scenarios.

We believe that \NAME{} can play an important role in guiding future methodological advances to overcome the demonstrated shortcomings of current saliency methods by systematically and quantitatively defining what correct behavior looks like on various tasks.
Moreover, \NAME{} can be extended over time as new methods are developed that perform consistently well on these stylized settings, by generating more sophisticated perception and reasoning tasks, e.g., by introducing semi-synthetic objects and/or encoding more complex reasoning through which the objects impact predictions. 
To facilitate this process, we provide source code\footnote{https://github.com/wnstlr/SMERF} that allows a user to (i) run the entire pipeline from generating datasets to computing results; 
 and (ii) evaluate new tasks by encoding new model reasoning and new methods.
Finally, generalizing the main ideas behind \NAME{} may also be useful in settings where saliency methods are inherently not appropriate, e.g. problems that require counterfactual model explanations~\cite{wachter2017counterfactual}.

\section*{Acknowledgements}

We thank Valerie Chen, Lucio Dery, Nari Johnson, and Mikhail Khodak for helpful feedback and discussions.
This work was supported in part by the National Science Foundation grants IIS1705121, IIS1838017, IIS2046613, IIS2112471, an Amazon Web Services Award, a Facebook Faculty Research Award, funding from Booz Allen Hamilton Inc., and a Block Center Grant. Any opinions, findings and conclusions or recommendations expressed in this material are those of the author(s) and do not necessarily reflect the views of any of these funding agencies.


\bibliography{main}
\bibliographystyle{icml2022}

\newpage
\appendix
\onecolumn

\section{Experiment Details}
\label{sec:appdx_training}

The repository of code used to generate the dataset and the results in the paper is available here\footnote{https://github.com/wnstlr/SMERF}.

\subsection{\texttt{TextBox} Dataset}
\label{sec:appdx_dataset}

The model for each reasoning was trained on a different number of training data points, based on the number of buckets available to the dataset, as shown below in Table~\ref{tab:number_training}. 
The total number of buckets depend on how the reasoning is set up (according to Table~\ref{tab:dataset_info}), i.e. how the labels are given based on the various combinations of features present/absent in the image. 
Based on this reasoning, primary and secondary objects are determined, which are objects in the image that should be highlighted (relevant for the prediction) and that should not be highlighted (not relevant for the prediction) respectively. 
If the object of interest is absent in the image, the corresponding primary/secondary metrics are not computed for further evaluations.

\begin{table*}[h]
    \centering
    \resizebox{\textwidth}{!}{\begin{tabular}{c c c c c c c c}
        Name & \textbf{Simple-FR} & \textbf{Simple-NR} & \textbf{Complex-FR} & \textbf{Complex-CR1} & \textbf{Complex-CR2} & \textbf{Complex-CR3} & \textbf{Complex-CR4} \\ \hline \hline
         Total Training & 24000 & 16000 & 36000 & 150000 & 150000 & 150000 & 150000\\ \hline
        Training/bucket & 2000 & 2000 & \begin{tabular}{@{}c@{}} 2000 (for 0) \\ 6000 (for 1)\end{tabular} & 15000 & 15000 & 15000 & 15000\\ \hline
        Total Validation & 6000 & 4000 & 6000 & 4000 & 4000 & 4000 & 4000\\ \hline
        Validation/bucket & 500 & 500 & 500 & 400 & 400 & 400 & 400\\ \hline
    \end{tabular}}
    \caption{The seven model reasoning settings considered in the experiments (Section~\ref{sec:experiments}) number of data points used.
    The total number of data points depend on a potential class imbalance between positive and negative samples, along with the total number of buckets.}
    \label{tab:number_training}
\end{table*}

Note that while all seven settings start with twelve buckets of images (since the same set of features is used throughout), some buckets may end up with undefined labels and will not be used to train the model.
For instance, Simple-NR (second column of Table~\ref{tab:dataset_info}) has 8 labeled buckets instead of 12, because images from four buckets without any Text (Buckets 1, 4, 7, 10 in Figure~\ref{fig:smerf_fr}) have undefined labels given that the specified model reasoning relies on Text to produce a label.
To balance the number of positive and negative samples in the dataset, more instances are sampled from buckets 10-12 for the training data for Complex-FR in particular. 
Finally, we note that even for a specified model reasoning, the object to focus on and/or avoid may differ across buckets. For instance, because Simple-NR relies only on Text for predictions, Text is always the object to focus for buckets in which it is present In contrast, Box1 and Box2 are objects to avoid in buckets when they appear, but of course cannot be avoided in buckets 2 and 3 where they do not appear in the first place.

\begin{figure}[!ht]
    \centering
    \includegraphics[width=0.7\textwidth]{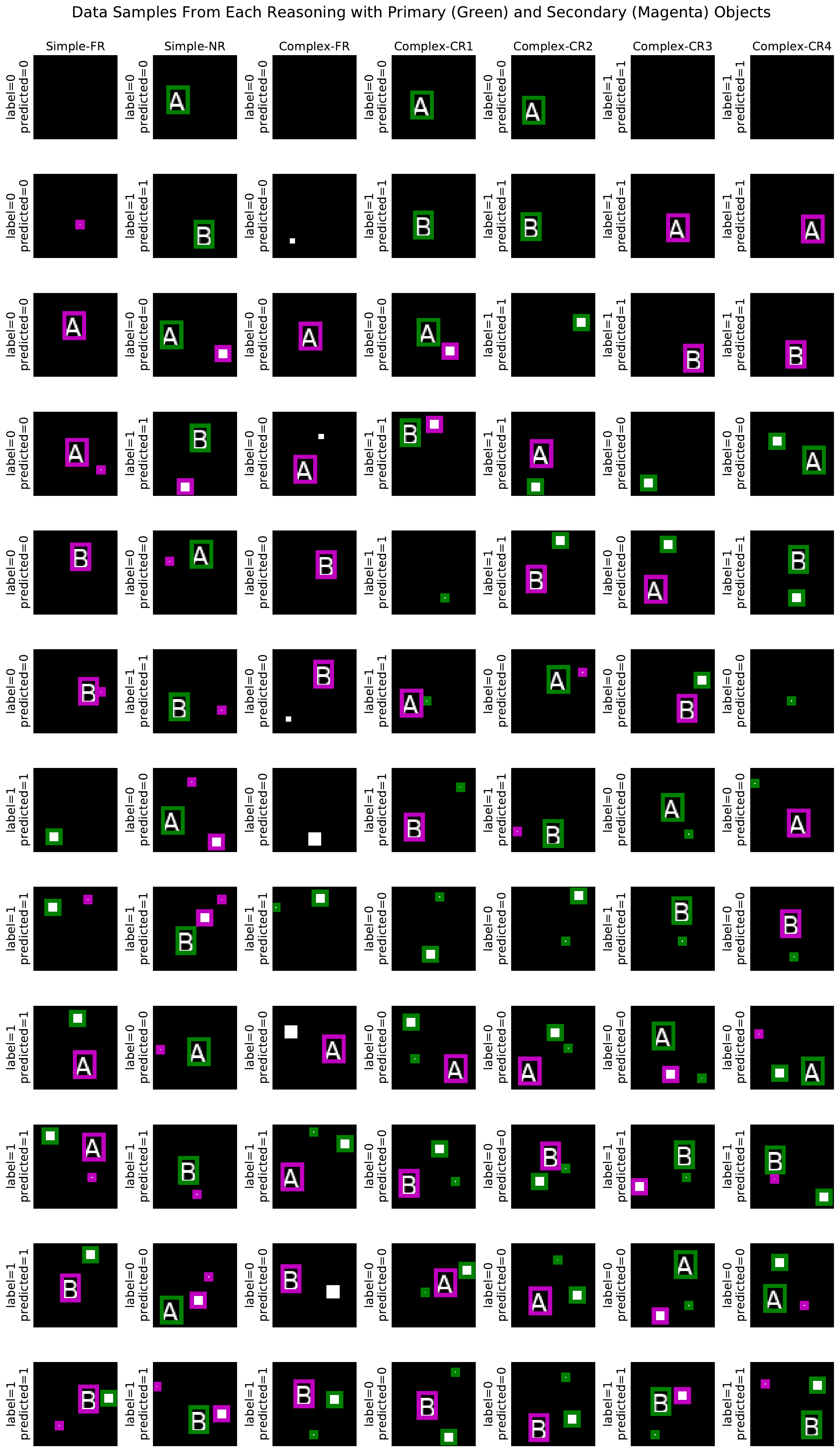}
    \caption{Data samples from each reasoning, with bounding boxes for primary (green) and secondary (magenta) objects, along with the trained model's predictions on them (all of which are correct).}
    \label{fig:datapoints}
\end{figure}

\texttt{TextBox} dataset is generated by sampling a random vector, each element indicating a specific feature the image should exhibit. The features include: the type of Text, location of the Text, color of the character, the background color, existence of Box1 and/or Box2. When randomly placing Box1 and/or Box2 on top of the Text images, the locations of these objects are constrained to avoid overlapping with one another. The resulting images have a dimension of 64-by-64 with three color channels. The pixel values are scaled to be between 0 and 1. In Figure~\ref{fig:datapoints}, we show some examples of data points for each reasoning, along with their labels, primary objects, and secondary objects. For each images, we also show what our trained model predicted (all of which are correct). The code for generating the dataset can be adapted to create different settings.

We show in Figure~\ref{fig:datapoints} some sample data points from different buckets with bounding boxes around primary and secondary objects for each reasoning. For each images, we also show what the trained model predicted (all of which are correct).

\newpage

\subsection{Model Architecture and Hyperparameters for Training}
\label{sec:appdx_model_training_hyperparameter}

For all results presented in the Experiments section (Section~\ref{sec:experiments}),  we trained separate convolutional neural networks for different types of model reasoning, but kept the same architecture of the same type. For simple reasoning, we have three convolutional layers (32 filters, kernel-size 3, stride (2,2); 64 filters, kernel-size 3, stride (2,2); 64 filters, kernel-size 3, stride (2,2)), followed by two fully-connected layers (200 units; 2 units), all with ReLU activation functions except for the output layer.
For complex reasoning, however, to account for the more complex feature relationships encoded in the dataset (lower number of parameters for the model could not achieve near-perfect accuracy on these datasets) the model has more parameters: four convolutional layers (64 filters, kernel-size 3, stride (2,2); 128 filters, kernel-size 3, stride (2,2); 256 filters, kernel-size 3, stride (2,2); 64 filters, kernel-size 3, stride (2,2)), followed by three fully-connected layers (200 units; 200 units; 2 units), all with ReLU activation functions except for the output layer. 

Learning rate was set as 0.0001, trained with Adam optimizer minimizing the binary cross entropy loss, with maximum epoch of 10. No particular grid search on these hyperparameters was performed, but the models were trained up to near-perfect accuracy for each model reasoning within several runs with different initializations (Table~\ref{tab:bucket_accuracy} shows bucket-wise test performance of the trained model for each reasoning). 
\NAME{} overall does not require much computational load as the model sizes are not big; the entire pipeline was tested out on a machine with a single GPU (GTX 1070, 8GB), with a system memory of 16GB. 

To further ensure that our findings are not affected by the specific choice of model architecture as described above, we additionally experimented with more complex model architectures: AlexNet~\cite{alexnet2012} and VGG16~\cite{simonyan2014very}. The results are presented in Appendix~\ref{sec:appdx_other_architectures}.

\begin{table}[h]
    \centering
    \small
    \resizebox{0.9\textwidth}{!}{\begin{tabular}{c c c c c c c c}
         & \textbf{Simple-FR} & \textbf{Simple-NR} & \textbf{Complex-FR} & \textbf{Complex-CR1} & \textbf{Complex-CR2} & \textbf{Complex-CR3} & \textbf{Complex-CR4} \\ \hline \hline
        Bucket1 & 1.00 & - & 1.00 & -  & - & 1.00 & 1.00\\ 
        Bucket2 & 1.00 & 1.00 & 1.00 & 1.00 & 1.00 & 1.00& 1.00\\ 
        Bucket3 & 1.00 & 1.00 & 1.00 & 1.00 & 1.00 & 1.00 & 1.00\\ 
        Bucket4 & 1.00 & - & 1.00 & 1.00 & - & - & 0.9975 \\ 
        Bucket5 & 1.00 & 1.00 & 1.00 & 0.95 & 1.00 & 0.9875 & 0.95\\ 
        Bucket6 & 1.00 & 1.00 & 1.00 & 1.00 & 0.9975 & 1.00 & 0.9225\\
        Bucket7 & 1.00 & - & 1.00 & - & 1.00 & 1.00 & - \\ 
        Bucket8 & 1.00 & 1.00 & 1.00 & 1.00 & 1.00 & 1.00 & 0.9975\\ 
        Bucket9 & 1.00 & 1.00 & 1.00 & 0.9575 & 1.00 & 0.995 & 0.9975\\ 
        Bucket10 & 1.00 & - & 1.00 & 1.00 & 1.00 & - & -\\ 
        Bucket11 & 1.00 & 1.00 & 0.968 & 1.00 & 0.975 & 1.00 & 1.00\\ 
        Bucket12 & 1.00 & 1.00 & 0.976 & 0.975 & 0.9125 & 0.9 & 0.995\\ 
    \end{tabular}}
    \caption{Bucket-wise test accuracy of the trained models.}
    \label{tab:bucket_accuracy}
\end{table} 

\subsection{Saliency Methods Tested}
\label{sec:appdx_saliency_methods}

We used various saliency methods as implemented in \texttt{iNNvestigate}\footnote{https://github.com/albermax/innvestigate} library~\cite{JMLR:v20:18-540}, which is licensed under the BSD License.
For methods that are not included in the library, we used the existing implementations for Grad-CAM\footnote{https://github.com/wawaku/grad-cam-keras} (Unlicensed) and DeepSHAP\footnote{https://github.com/slundberg/shap} (MIT Licensed) and integrated them into the pipeline. 
The details of the hyperparameters used for these methods are all available in the source code, specifically in \texttt{smerf/explanations.py} file. No particular grid search was performed for these hyperparameters. 

\newpage

\subsection{Evaluation Metric: Intersection-Over-Union (IOU) and Attribution Focus Level (AFL)}
\label{sec:appdx_eval_metric_afl}

Instead of using the typical IOU value as our main metric\footnote{Nevertheless, we provide full results on IOU in Appendix~\ref{sec:appdx_metric_results}, which are consistent with our results with AFL.}, we use Attribution Focus Level (AFL), which alternatively computes the proportion of attribution values assigned to specific objects out of the total attribution value (normalized to be 1).
Below we will demonstrate the advantage of using AFL over IOU in the context of \NAME{}.

\subsubsection{IOU loses information from raw feature attribution values, so misses the impact of signals from other objects in the image.}

\begin{figure}[!ht]
    \centering
    \includegraphics[width=\textwidth]{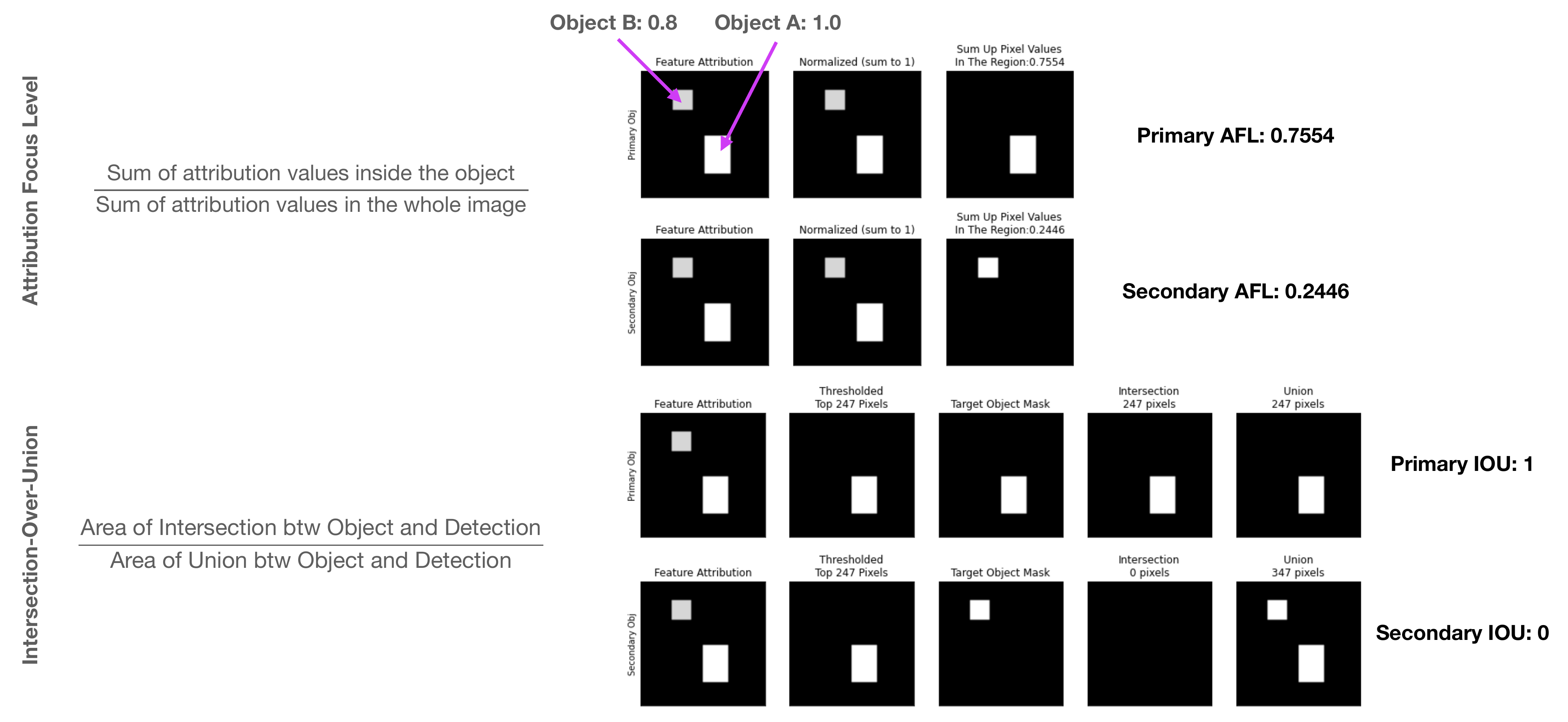}
    \caption{Computing AFL vs IOU on a toy example where attribution values from a secondary object are non-trivial. In this example, object A (primary) and B (secondary) are both highlighted by the feature attribution as shown, where the attribution values for each pixel of A being 1, B being 0.8, and background being 0.001. AFL is computed by taking the sum of values inside the object of interest and dividing that to the sum of attribution values in the whole image. This results in PAFL value of 0.75, and SAFL value of 0.24. Notice the non-zero SAFL that indicates non-trivial signal from the secondary object B. On the other hand, typical IOU, due to its thresholding to contain only a limited number of pixels for the evaluation, fails to capture this non-trivial signal from object B, resulting in the Secondary IOU value of 0.}
    \label{fig:metric_example}
\end{figure}

Consider a feature attribution which focuses on two objects A and B in an image (A being primary and B being secondary), where the attribution values of pixels inside object A are all 1, those inside object B are all 0.8, and 0.001 elsewhere in the background. The example is shown in Figure~\ref{fig:metric_example}. PAFL and SAFL for this feature attribution are 0.75 and 0.24 respectively. Note that the SAFL value is non-zero, correctly reflectin the fact that the feature attribution values are non-trivial on the secondary object despite most of the attribution still focused on the primary object. 

However, IOU fails to correctly show this non-trivial signal on the secondary object (ie.e Secondary IOU is zero). This is mainly due to the thresholding process which selects only top-K pixels with the highest attribution values and masking only those to be included when computing the areas of intersection and union. For this example, as shown in the image on the second column of Figure~\ref{fig:metric_example} for IOU, 247 pixels are selected from thresholding, which ends up selecting only the region of object A, as the pixels within this region all have higher attribution values (1) compared to those in object B (0.8). This results in Primary IOU and Secondary IOU of 1 and 0 respectively, which fails to capture non-trivial strength of focus on the secondary object. As \NAME{}'s model reasoning settings require analyzing the feature attributions' strength on both primary and secondary objects, it is important to correctly address this issue. 
\subsubsection{IOU is sensitive to the number of pixels to include when thresholding.}

\begin{figure}[t]
    \centering
    \includegraphics[width=0.9\textwidth]{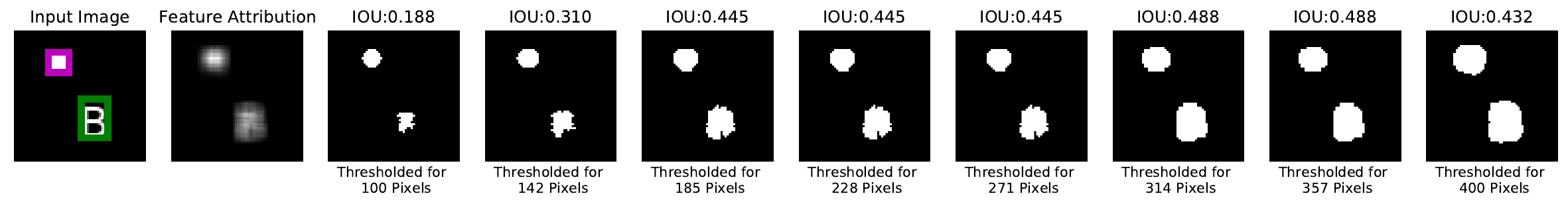}
    \caption{Change of thresholded masks based on the number of pixels to include, and the corresponding Primary IOU values on feature attributions obtained with Integrated Gradients. The IOU values are sensitive to the amount of thresholding.}
    \label{fig:iou_change}
\end{figure}

Another problem with thresholding is that it is usually unclear what the thresholding value should be. In the above example, 247 pixels are selected to include the number of pixels that consists the primary object. While this is an intuitive choice, this value can be arbitrary and the resulting IOU values can vary significantly based on the choice. Figure~\ref{fig:iou_change} shows Primary IOU from a feature attribution obtained by Integrated Gradients on one of the data points where the Text is primary and the Box1 is secondary. 

\subsubsection{AFL provides more intuitive understanding about the saliency methods' level of focus among various objects.}

\begin{figure}
    \centering
    \includegraphics[width=\textwidth]{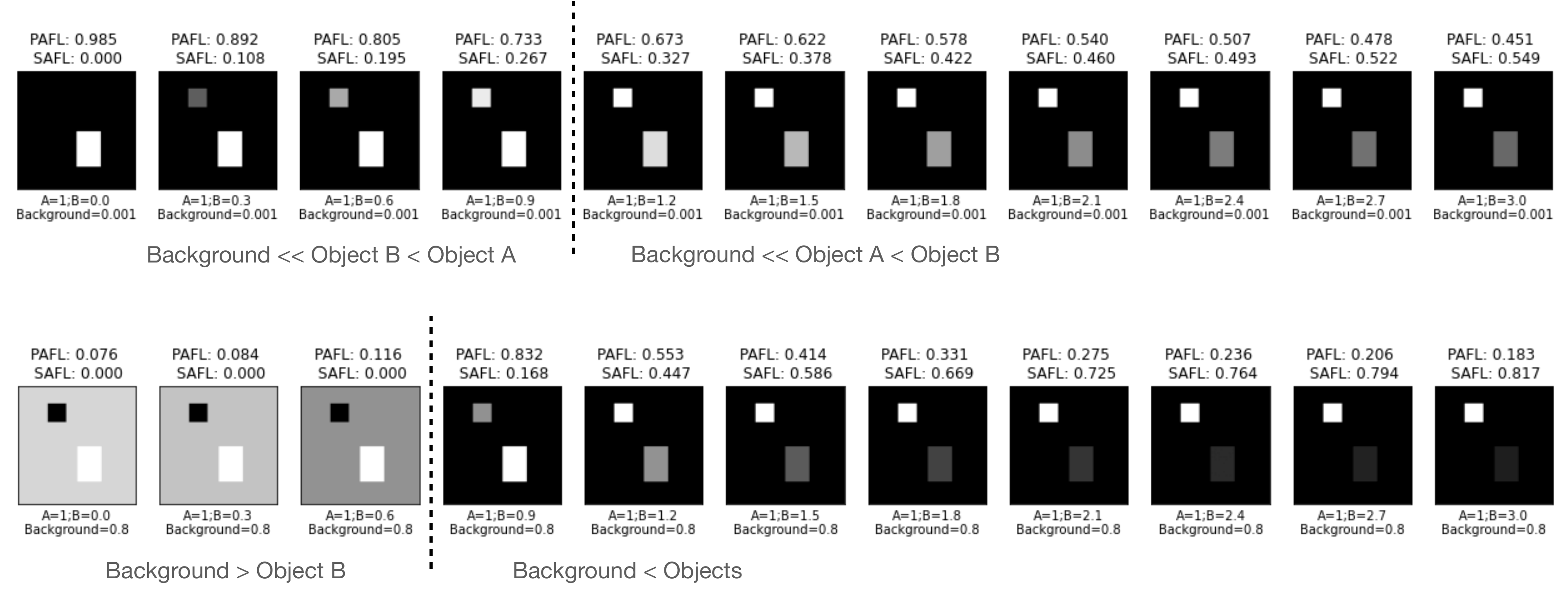}
    \caption{Taking the example of Objects A/B from Figure~\ref{fig:metric_example}, (Top) as the strength of attribution values placed on object B increases, the PAFL values decrease and SAFL values increase. With the background attribution values being significantly smaller than the objects, PAFL and SAFL sums up to one. Naturally, when SAFL $>$ PAFL, it more focus is attributed to the secondary object compared to the primmary. (Bottom) When the attribution values on the background are high (higher than objects), PAFL and SAFL approaches zero as the background dominates the share in the total attribution (first three images). So when both PAFL and SAFL is low, it represents the case where most of the signals are coming from the background.}
    \label{fig:afl_primary_secondary}
\end{figure}

The sum of PAFL and SAFL is upper-bounded by one, as the sum of all attribution values in the image is normalized to sum up to 1. By comparing these values, it is intuitive to deduce whether more focus is present on the primary/secondary object, or on the background. Also it is easy to understand how the level of focus shifts among these objects.  Figure~\ref{fig:afl_primary_secondary} shows an example of how different degrees of attribution values placed on the secondary object (top) and the background (bottom) can change the corresponding PAFL and SAFL values. 
Note that as more we can observe PAFL decreasing and SAFL increasing, as more proportion of the total attributions are assigned to the secondary object (top). Also when the background has higher attribution values than the objects, both PAFL and SAFL values plummet. 



\newpage

\section{Additional Experimental Results}
\label{sec:appdx_exp_result}

\subsection{Results with IOU Metric}
\label{sec:appdx_metric_results}

Intersection-Over-Union (IOU) metric is a ratio of the intersecting area to the union area of the 0-1 masked feature attribution and the ground-truth feature attribution. 
The ground-truth feature attributions are predefined from the data generation process (by the ground-truth model reasoning). 
The 0-1 masked feature attribution from the saliency methods is obtained by first blurring the original feature attribution averaged across the three color channels, followed by thresholding the pixel intensity to select top-$K$ pixels, where $K$ is equal to the number of pixels that correspond to the primary object.  
Given the 0-1 masked feature attribution and the ground-truth feature attribution, we compute two types of IOU values just like AFL: 
(1) primary IOU (PIOU), which measures how much of the 0-1 masked feature attribution overlaps with the ground-truth for the region \textit{relevant} to the model prediction; 
and (2) secondary IOU (SIOU), which measures the same value with respect to the region \emph{not} \textit{relevant} to the model prediction. 
Just like AFLs, PIOU should be high and SIOU should be low for methods that are more effective and correct. Throughout this section we use the threshold value of 0.5 to roughly distinguish good and bad performance in terms of IOU as commonly done in practice~\cite{everingham2015pascal, wang2019pseudo}. 

\begin{figure}[h]
    \centering
     \begin{subfigure}[b]{0.62\textwidth}
         \centering
          \includegraphics[width=\textwidth]{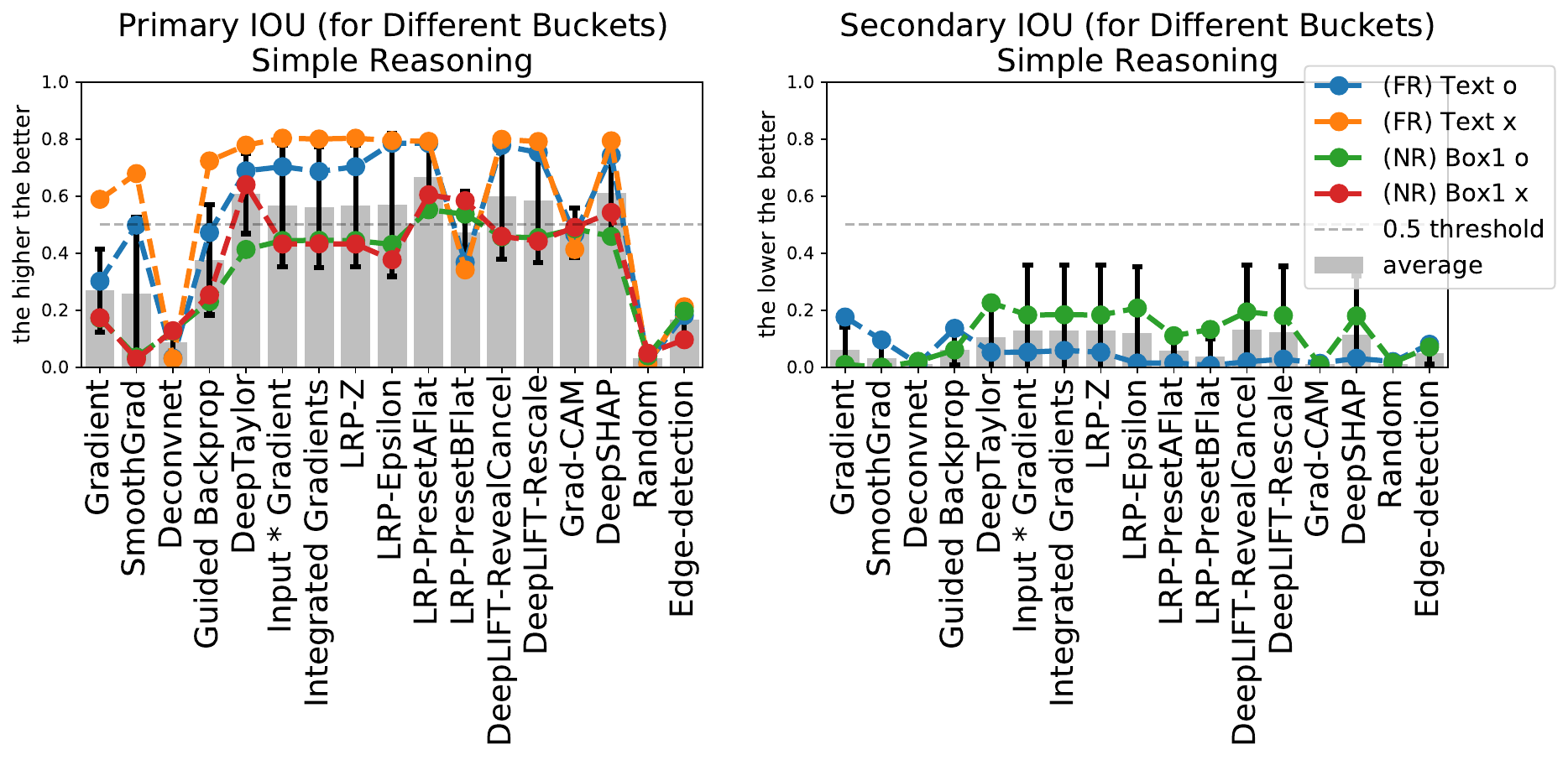}
         \caption{PIOU and SIOU for Simple Reasoning}
         \label{fig:simple_ious}
     \end{subfigure} 
     \hfill
     \begin{subfigure}[b]{0.32\textwidth}
         \centering
          \includegraphics[width=\textwidth]{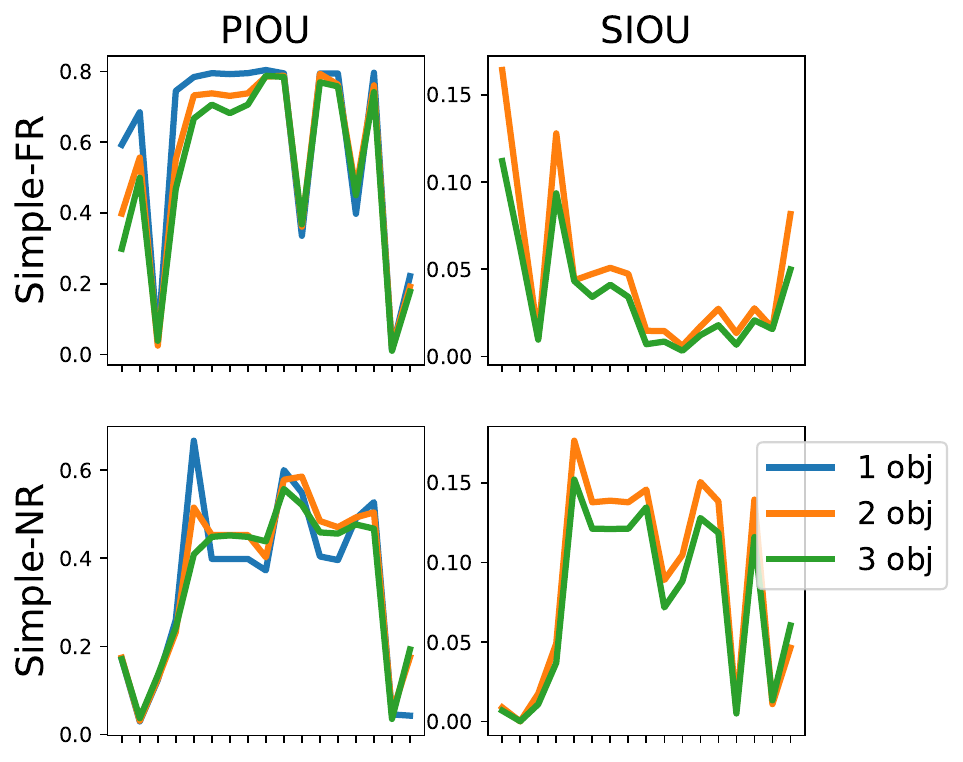}
         \caption{PIOU and SIOU based on the number of objects in the image (x-axis is the same as Figure~\ref{fig:simple_ious})}
         \label{fig:simple_ious_obj}
     \end{subfigure}
    \caption{(a) PIOU (left) and SIOU (right) for Simple-FR and Simple-NR from Table~\ref{tab:dataset_info}. 
    The black vertical lines indicate the standard deviation across different buckets. 
    The dotted horizontal line is the correctness threshold of 0.5. 
    Most methods perform well on average, with reasonable PIOU and low SIOU, except for a few methods on the left.
    (b) There is not much difference between the PIOU and SIOU values based on the number of objects in the image. 
    }
    \label{fig:simple_result_iou}
\end{figure} 

\begin{figure}[h]
    \begin{subfigure}[b]{0.72\textwidth}
         \centering
         \includegraphics[width=\textwidth]{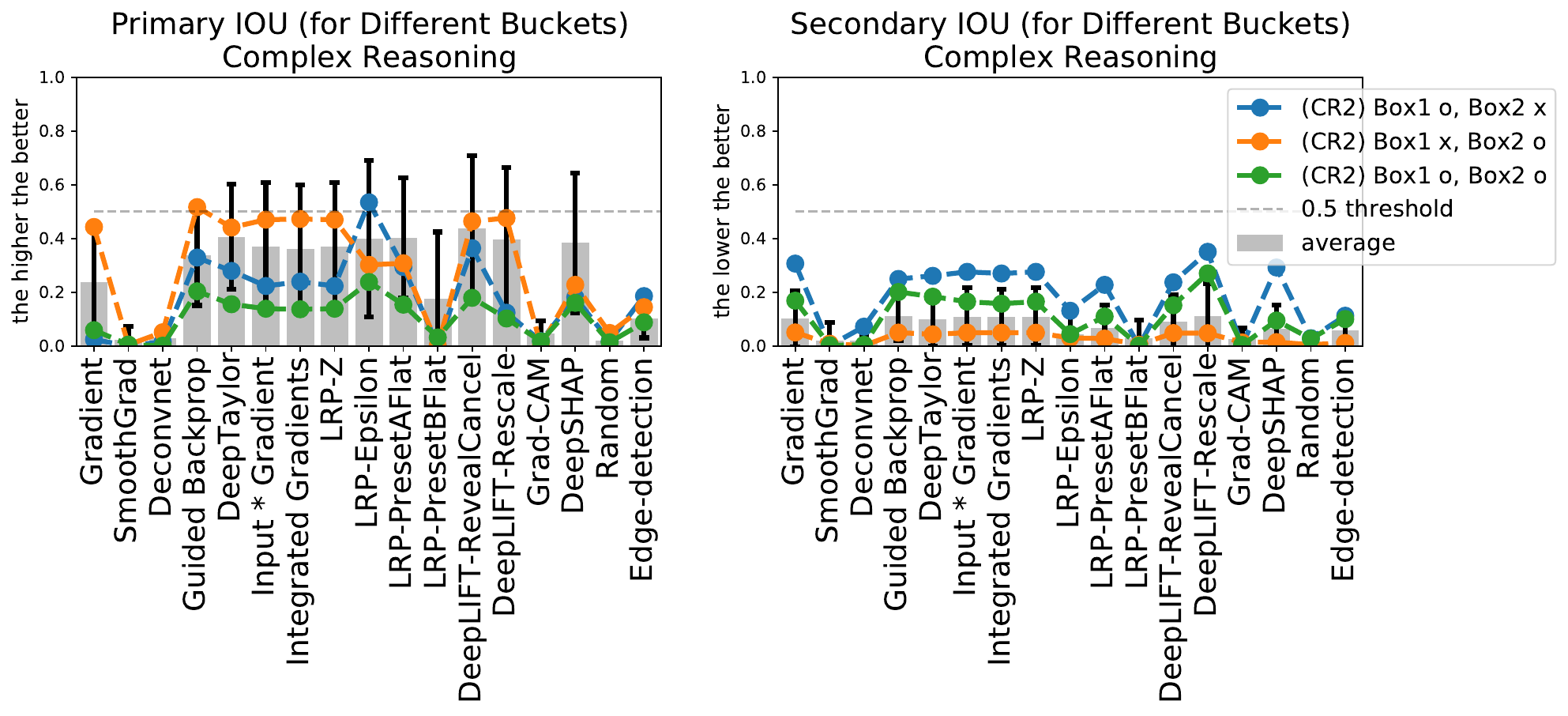}
         \caption{PIOU and SIOU for Complex Reasoning}
         \label{fig:complex_ious}
     \end{subfigure} 
    \hfill
    \begin{subfigure}[b]{0.27\textwidth}
         \centering
         \includegraphics[width=\textwidth]{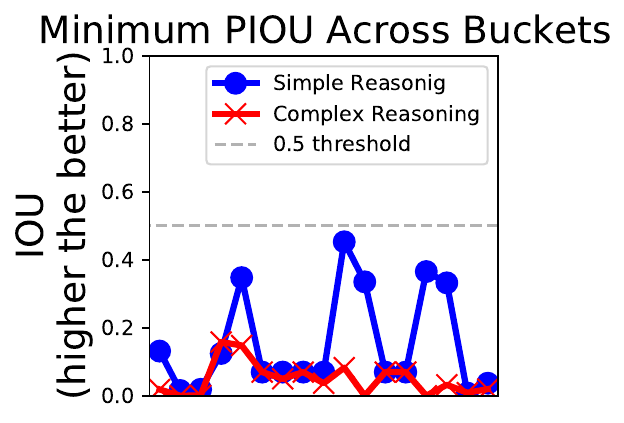}
         \caption{Minimum PIOU across buckets on simple (blue circle) vs. complex (red x) reasoning (x-axis is the same as Figure~\ref{fig:complex_ious}).}
         \label{fig:complex_iou_min}
     \end{subfigure} 
    \caption{(a) PIOU and SIOU for complex reasoning. 
    Average performance on PIOU is mostly lower than 0.5, with worse performance compared to the results from simple reasoning. 
    (b) Worst-case buckets show worse PIOU values in complex reasoning. 
    }
    \label{fig:complex_result_iou}
\end{figure} 

Figure~\ref{fig:simple_ious} shows PIOU and SIOU values for simple reasoning settings. 
Note that most methods achieve PIOU higher than 0.5, just like the results from AFL.
SIOU values are mostly low throughout, even lower than what we observed from AFL. However, we notice that the IOU values fail to capture the unexpected variation of level of focus given to multiple  objects based on how saturated the images are, unlike what we observed from AFLs (Figure~\ref{fig:simple_ious_obj}).
 
Figure~\ref{fig:complex_ious} shows the PIOU and SIOU values for complex reasoning settings. 
In this case, all methods score PIOU below 0.5 correctness threshold, just like AFL. Also the minimum PIOU across different buckets from complex reasoning is strictly lower than those from simple reasoning for all methods (Figure~\ref{fig:complex_iou_min}). Therefore, general trend of methods being reasonably good for simple reasoning and being bad for complex reasoning still is observable using IOU.
However, note that the SIOU values remain relatively low for both cases despite the decrease in PIOU for complex reasoning. 
This is where the aforementioned limitation of IOU metric is exposed, where it is difficult to clearly deduce where the changes in the primary values are coming from, contrary to what we observe from AFL.

Relatedly, IOU metric is relatively more generous to primary objects due to the way they are thresholded and it is more prone to ignoring potential non-trivial signals from objects that are not primary. 
This naturally leads to low SIOU values and a lack of clear relationship between primary and secondary metric: an increase in one would not necessarily imply a decrease in another, and vice-versa. 
In addition, the loss of information from the attribution values makes it difficult to better understand why one value decreased and how much of that change could be attributed to different objects in the image. Due to these limitations (as well as examples mentioned earlier in Appendix~\ref{sec:appdx_eval_metric_afl}, we present AFL results in the main text for better insights and highlighting the shortcomings of the methods in different settings.  


\subsection{Qualitative Results}
\label{sec:appdx_qual_results}

\begin{figure}
    \centering
    \includegraphics[width=\textwidth]{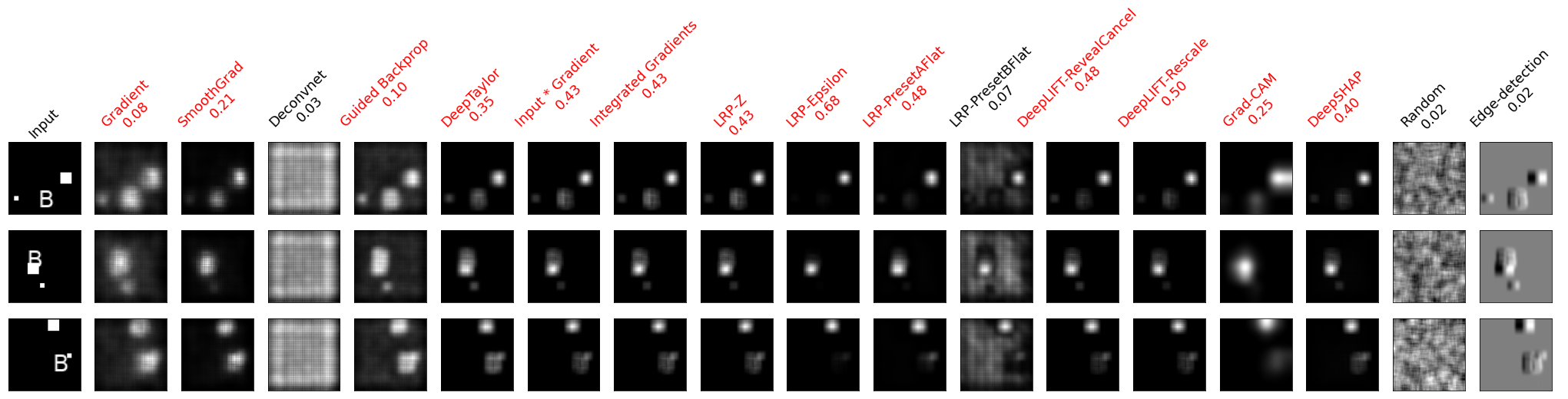}
    \includegraphics[width=\textwidth]{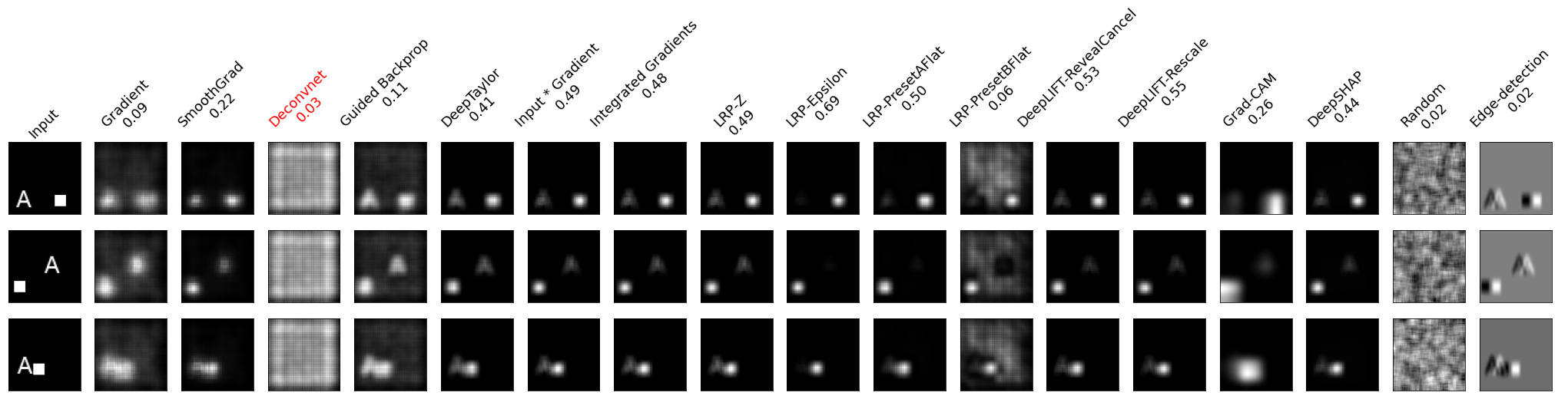}
    \includegraphics[width=\textwidth]{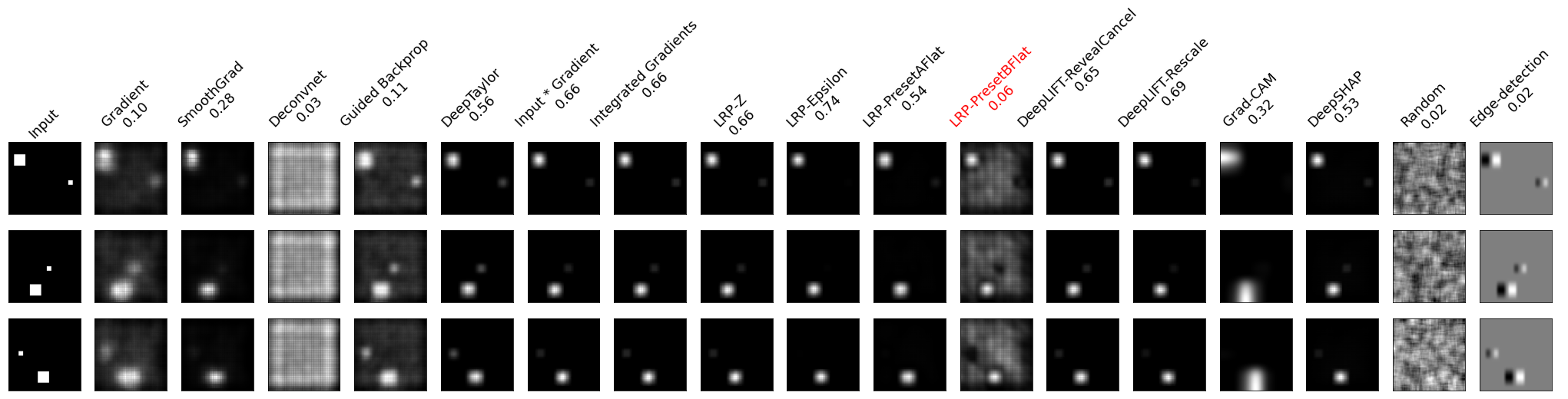}
    \caption{Image samples from buckets where minimum PAFL values are recorded for different methods (marked with red), for FR-Simple. The ground-truth is to focus on Box1 only.}
    \label{fig:FR_fails}
\end{figure}

\begin{figure}
    \centering
    \includegraphics[width=\textwidth]{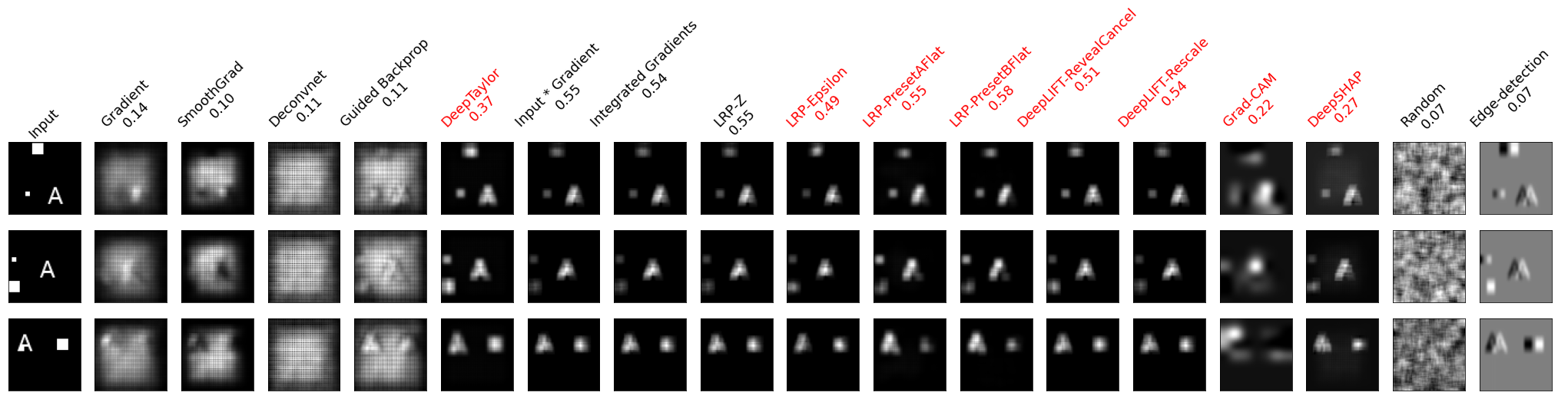}
    \includegraphics[width=\textwidth]{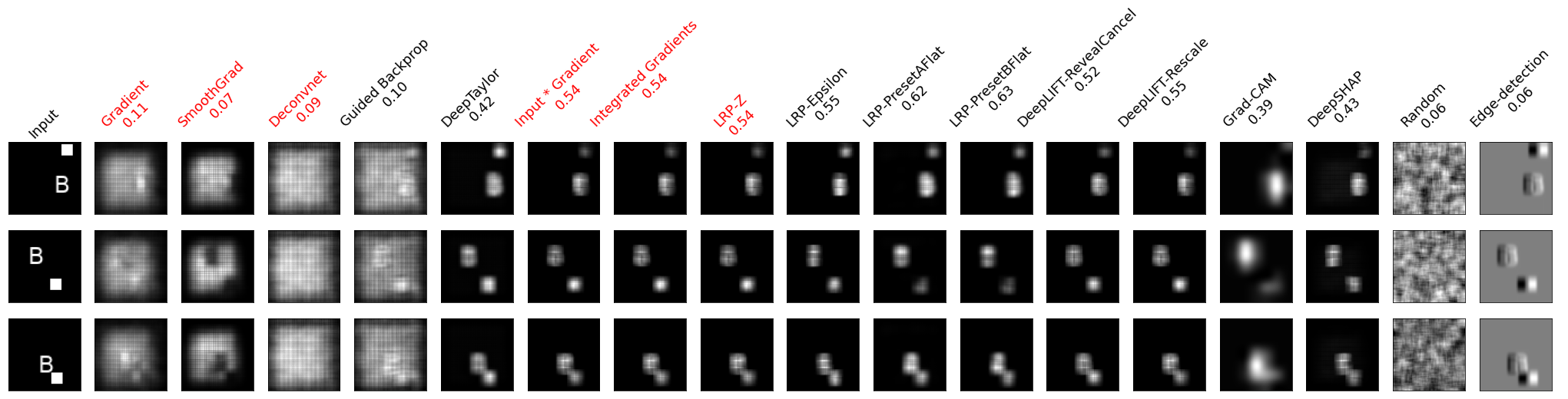}
    \includegraphics[width=\textwidth]{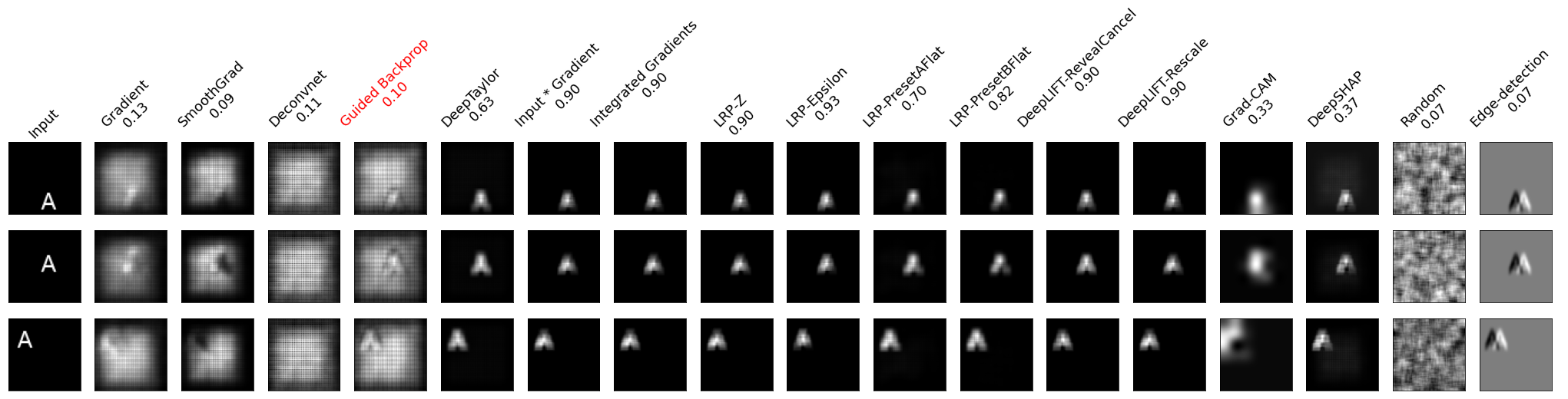}
    \caption{Image samples from buckets where minimum PAFL values are recorded for different methods (marked with red, along with corresponding PAFL values), for NR-Simple. The ground-truth is to focus on the Text only.}
    \label{fig:NR_fails}
\end{figure}

Figure~\ref{fig:FR_fails} and ~\ref{fig:NR_fails} show samples from simple reasoning (each from Simple-FR and Simple-NR) in buckets where the methods record the lowest PAFL (indicated with red). For instance, in Simple-FR setup (Figure~\ref{fig:FR_fails}) almost all methods achieve the lowest PAFL on the bucket where all features are present (top panel). Recall that Simple-FR relies fully on Box1; however we observe that several methods highlight all objects that are present in the image to a certain degree. Such tendency explains the unexpected variation of AFL based on the number of objects present in the image. Nevertheless, some minimum PAFL values are still above 0.5 for a method like LRP-Epsilon, which makes it the most effective method for Simple-FR that is less prone to errors (it also correctly focuses on Box1 without being distracted by other objects in the image). Compared to Simple-FR, Simple-NR shows more evenly distributed failure cases across different buckets shown (Figure~\ref{fig:NR_fails}).  

\begin{figure}
    \centering
    \includegraphics[width=\textwidth]{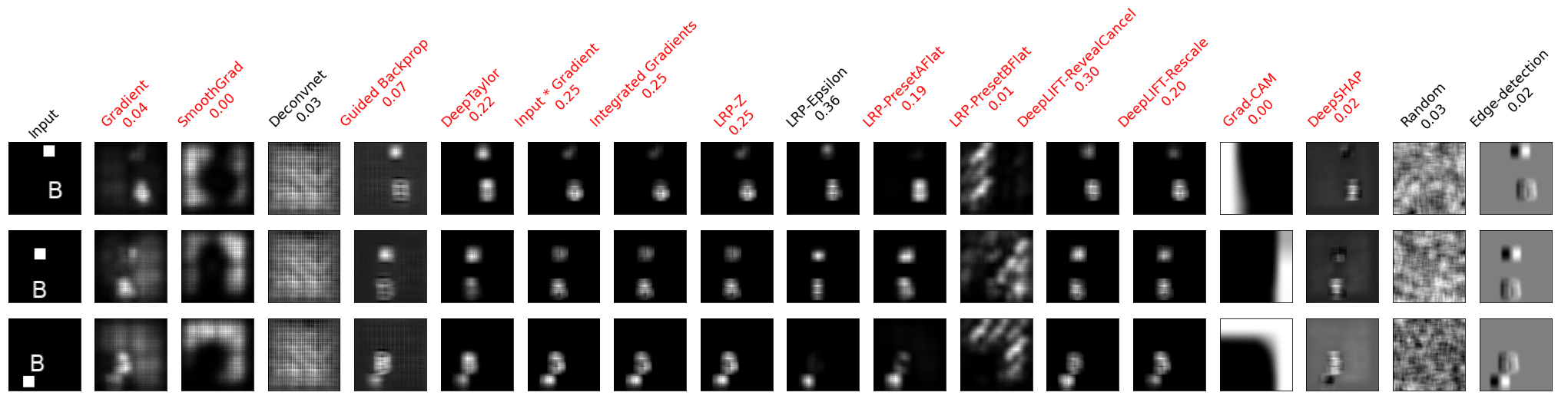}
    \includegraphics[width=\textwidth]{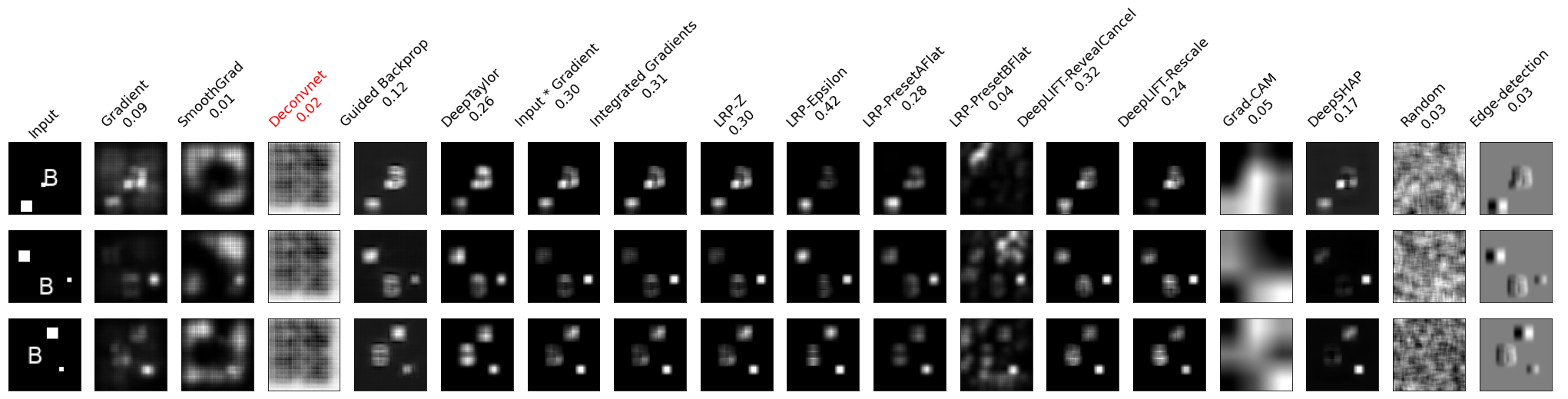}
    \includegraphics[width=\textwidth]{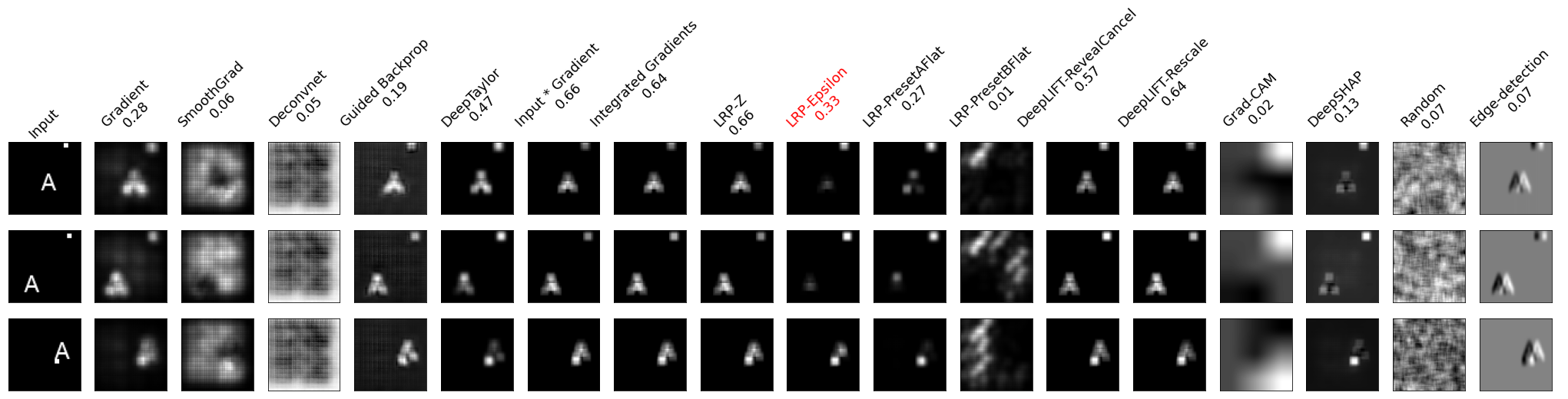}
    \caption{Image samples from buckets where minimum PAFL values are recorded for different methods (marked with red), for CR-Complex2. The ground-truth is to focus on the Box1 and Box2 if Box1 is present, otherwise on Text.}
    \label{fig:CR_fails}
\end{figure}

In Figure~\ref{fig:CR_fails} we have the samples from buckets with minimum PAFL for complex reasoning setup, in particular, Complex-CR2. This time most of the cases with minimum performance is concentrated on the bucket with Box1 and Text B (top panel; 13 out of 15 methods have minimum PAFL on this bucket). Recall that for this bucket, the ground-truth feature attribution should highlight just Box1. For most of the methods, more focus is given to the Text, which results in a particularly low PAFL across all methods. Notice that all the values are below 0.5: this aligns with our earlier observation that the worst-case performance of the methods on complex reasoning setup is much worse than simple reasoning setup. 

\newpage

\subsection{Per-bucket Performance Variation for Complex Reasoning}
\label{sec:appdx_complex_reasoning_perbucket}

\begin{figure}[h]
    \centering
    \includegraphics[width=0.52\textwidth]{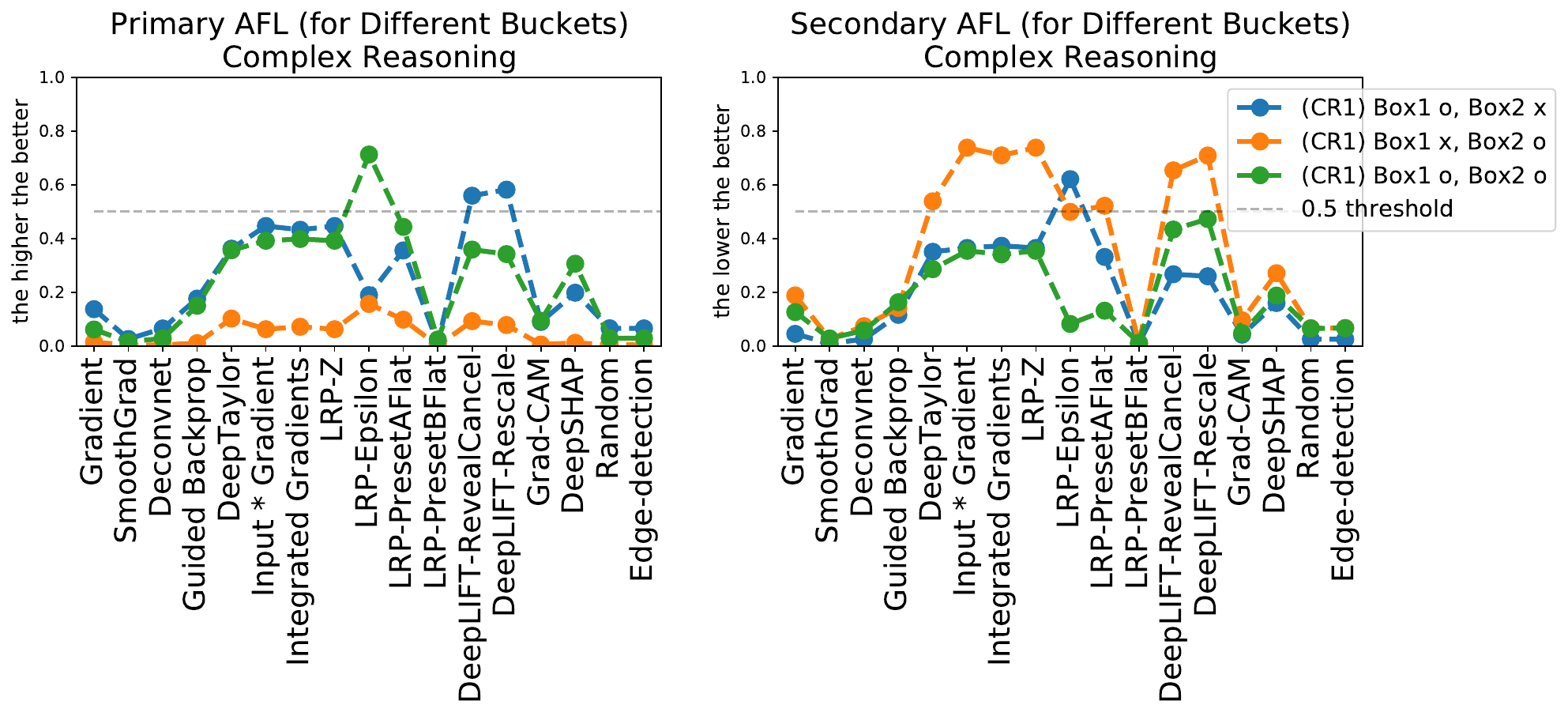}
    \includegraphics[width=0.52\textwidth]{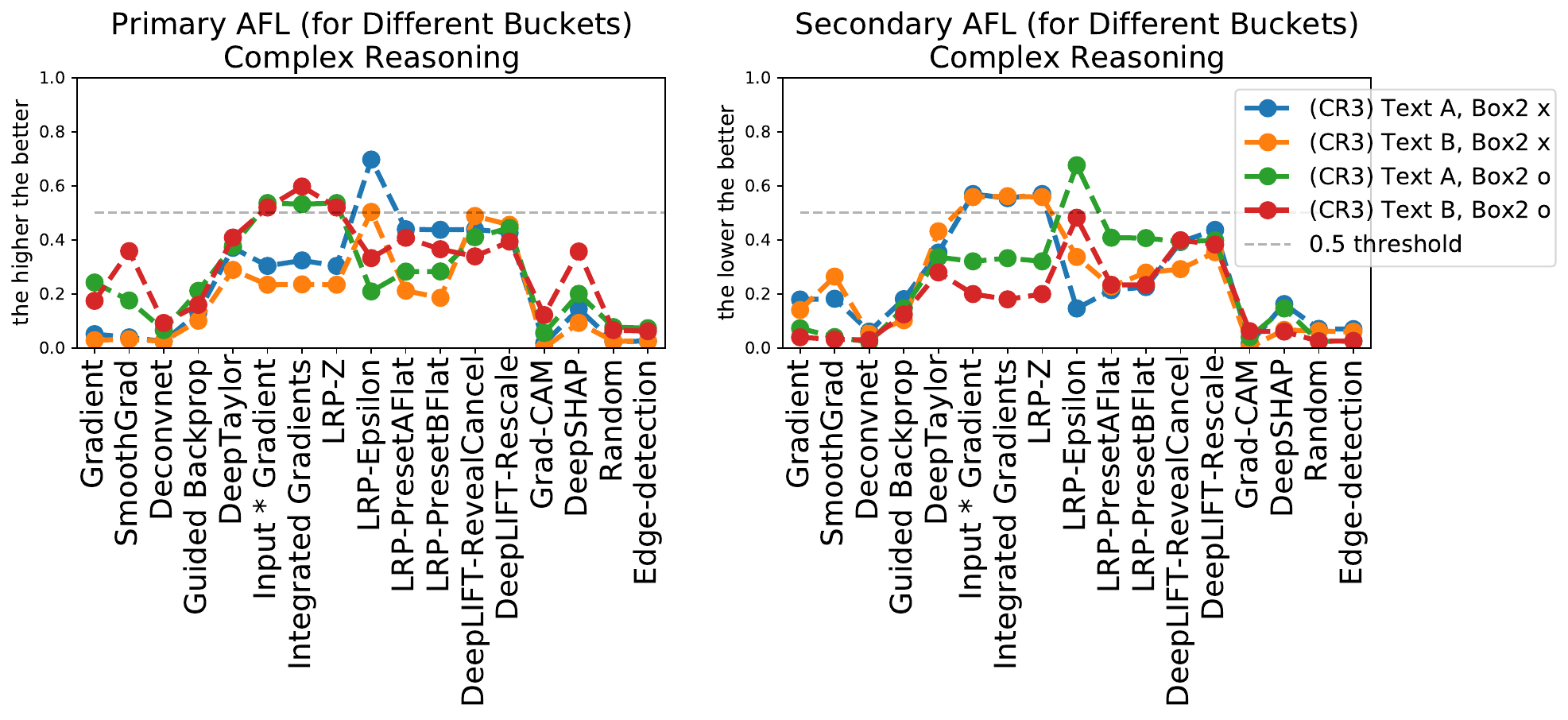}
    \includegraphics[width=0.52\textwidth]{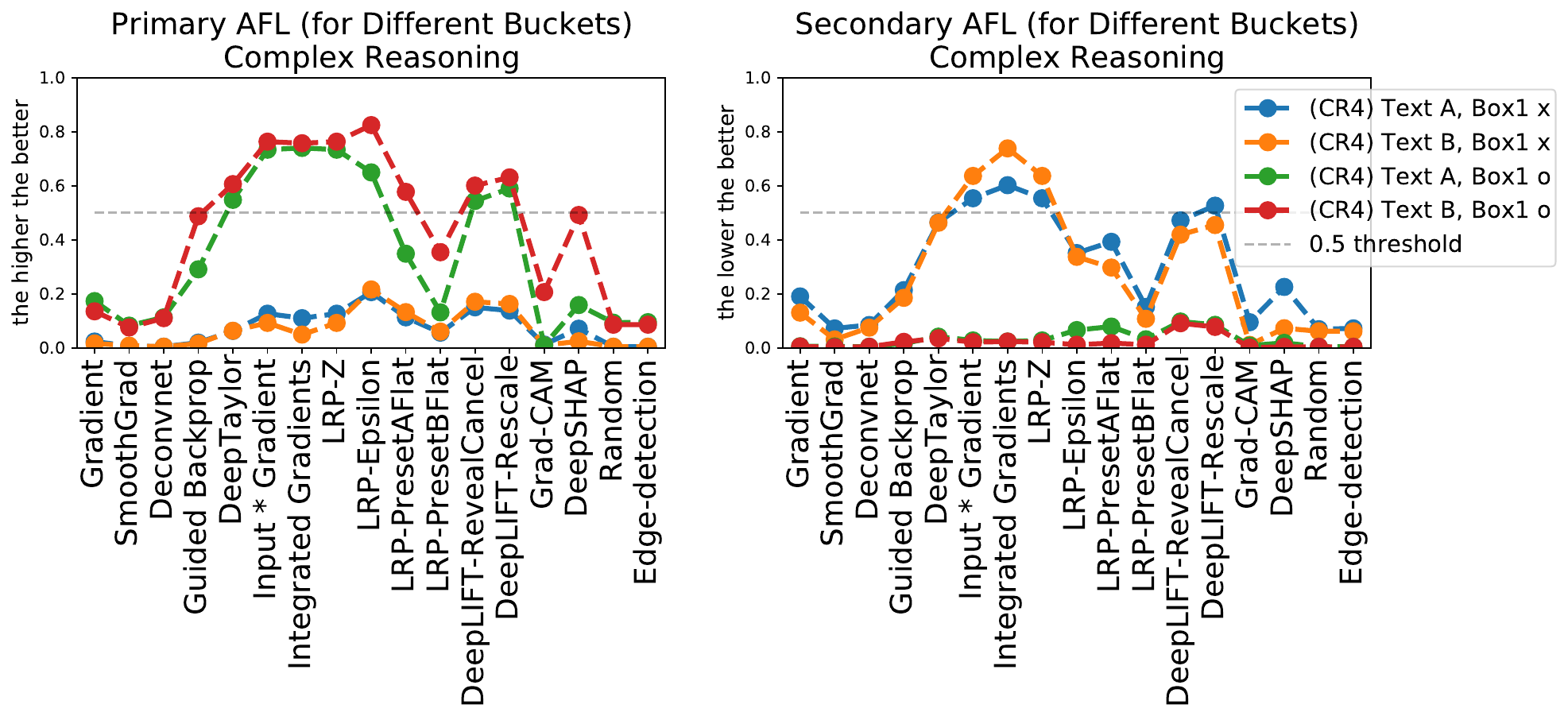}
    \includegraphics[width=0.52\textwidth]{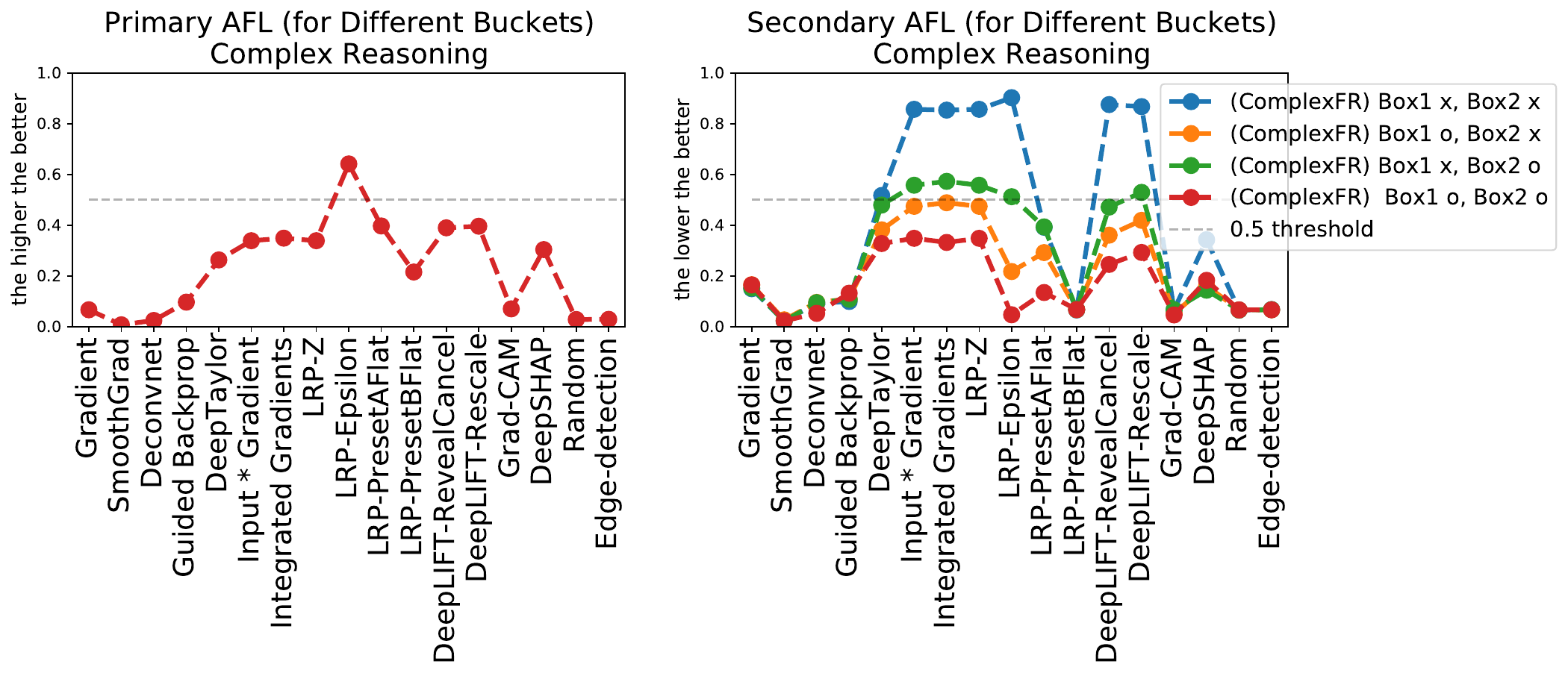}
    \caption{Additional information for the variance of PAFL and SAFL across different buckets for complex reasoning (from top to bottom, Complex-CR1, Complex-CR3, Complex-CR4, Complex-FR). The performance varies significantly from bucket to bucket, just as we described in the main text.}
    \label{fig:CR_perbucketvar}
\end{figure}

In Figure~\ref{fig:CR_perbucketvar}, we show additional details from different types of complex reasoning that show high variance of PAFL and SAFL we observed from Figure~\ref{fig:complex_afls}. 
From top to bottom, we plot the variation of the values per bucket for Complex-CR1, Complex-CR3, Complex-CR4, and Complex-FR, in addition to the results from Complex-CR2 reported in the main text. 
We similarly observe that the variation is quite extreme for different buckets. 

\newpage

\subsection{Additional Failure Case Analysis}
\label{sec:appdx_failure_cases}

While the AFL may be more suited to give a high-level understanding of the methods' performance, to further understand the degree to which the model focuses more on the incorrect region compared to the correct region (i.e. to identify failure due to wrong focus), we also compute the mean of the attribution values inside the relevant (which we call \textit{Primary Mean-AFL, or PMAFL}) and irrelevant (which we call \textit{Secondary Mean-AFL, or SMAFL}) regions for comparison. 
For a successful method, primary MAFL should always upper bound secondary MAFL because on average the correct regions should always be assigned higher attribution values compared to the incorrect regions. 
In this analysis, the actual values of primary and secondary MAFL do not matter; only a relative comparison does.

\begin{figure}[ht]
    \centering
    \includegraphics[width=\textwidth]{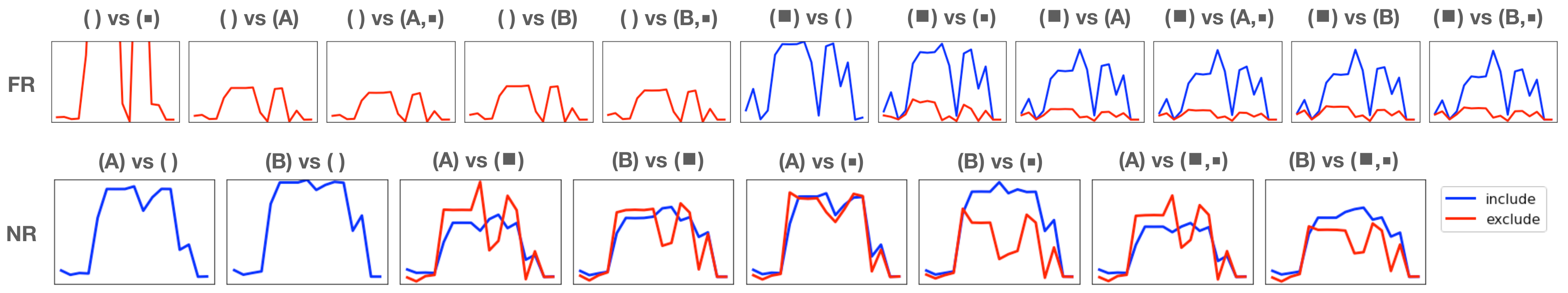}
    \caption{MAFL value comparison within buckets for simple reasoning with FR and NR. Title of each panel indicates the bucket with ``( correct feature ) vs ( incorrect feature )'', and the blue and red lines show the general trend of primary and secondary MAFL values across different methods (x-axis being the same as Figure~\ref{fig:simple_afls}, omitted for brevity). Primary MAFL should ideally always upper-bound secondary MAFL: while that is the case for FR, there are occasional failure cases in NR where the opposite happens.}
    \label{fig:avg_simple}
\end{figure}

\begin{figure}[ht]
    \centering
    \includegraphics[width=\textwidth]{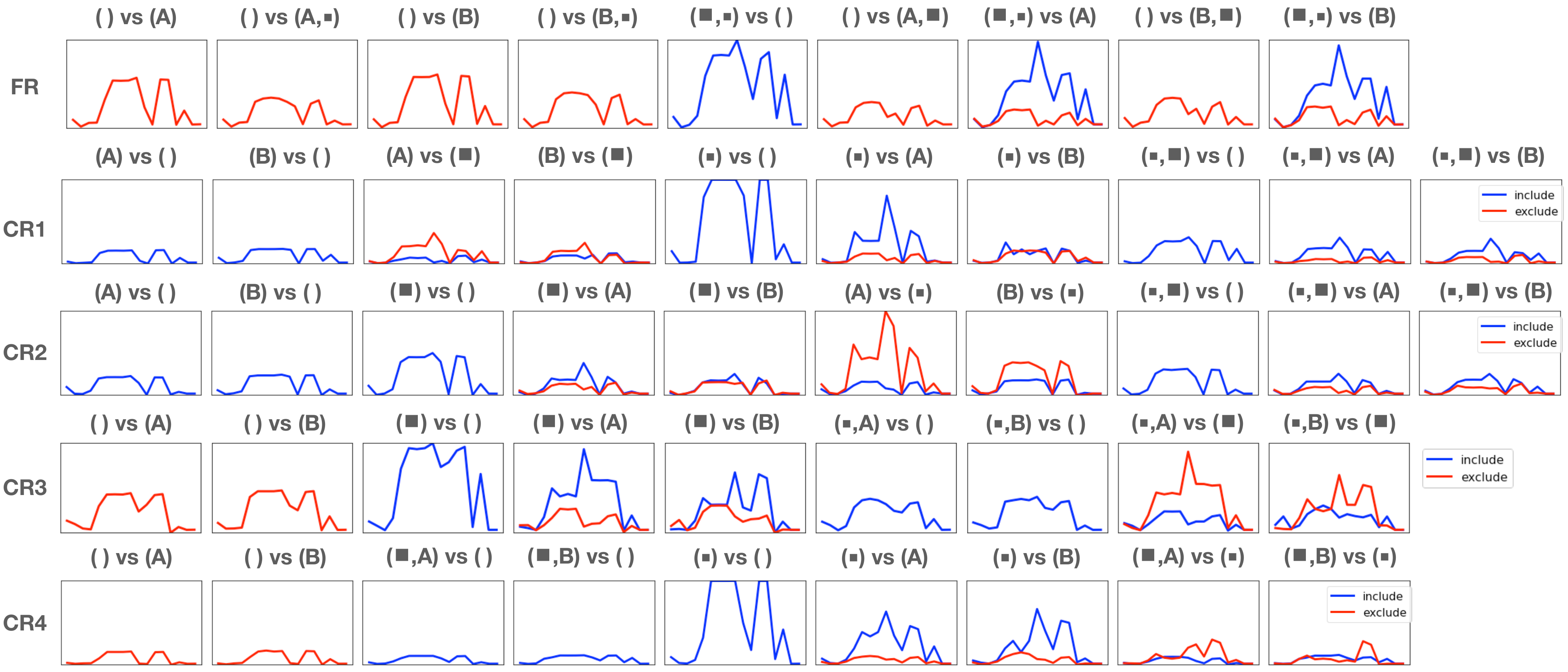}
    \caption{MAFL values for complex reasoning. While PMAFL (blue) should be always bigger than SMAFL (red), some buckets exhibit much larger SMAFL. This means the methods are failing due to focusing significantly more on irrelevant regions features than relevant ones.}
    \label{fig:avg_complex}
\end{figure}

Figure~\ref{fig:avg_simple} plots MAFL value comparison for different buckets in Simple-NR and Simple-FR. It is important that the blue line (PMAFL) should always upper-bound the red line (SMAFL). While that is the case for FR for all buckets, there are occasional cases in NR with the opposite relationship for certain methods. Although the gap between the values are not big, this still indicates that the feature attribution may sometimes be wrongly focusing on irrelevant regions of the image more than the correct ones. Such failure cases have not been actively discussed in previous works that dealt models with simple reasoning~\cite{yang2019benchmarking, adebayo2020debugging}. 

As further shown in Figure~\ref{fig:avg_complex} it is observed that the gap between PMAFL (blue) and SMAFL (red) is bigger for some buckets in complex reasoning compared to the simple reasoning case in Figure~\ref{fig:avg_simple}. 
Such larger gaps clearly demonstrate and verify the methods' more frequent failure due to wrong focus from complex reasoning. 

\newpage

\subsection{Identifiability Problem Samples}
\label{sec:appdx_identify}

\begin{figure}[ht]
    \centering
    \includegraphics[width=0.60\textwidth]{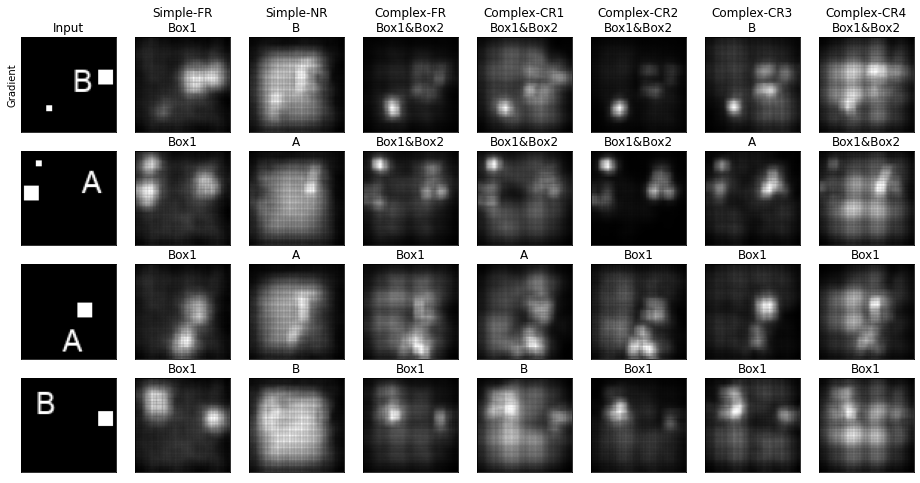}
    \includegraphics[width=0.60\textwidth]{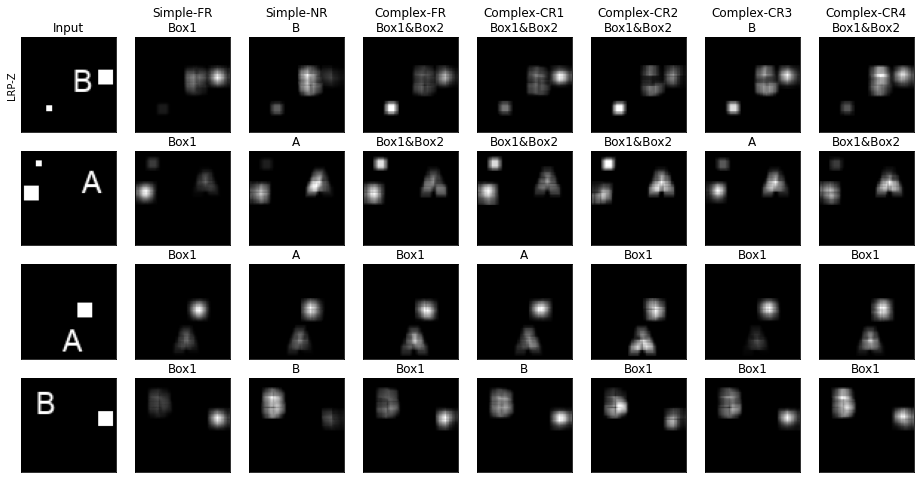}
    \includegraphics[width=0.60\textwidth]{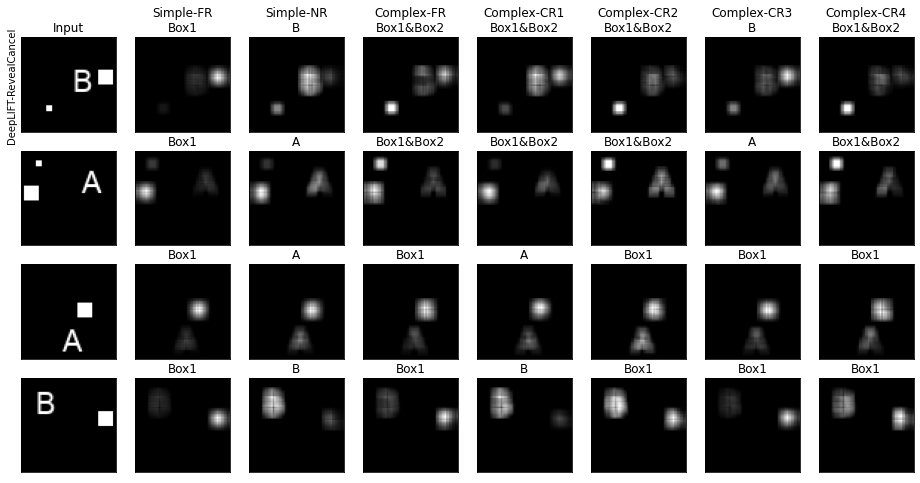}
    \caption{Identifiability problem samples with other methods (Gradient, LRP-Z, DeepLIFT from top to bottom). All methods highlight all objects in the image up to a certain degree regardless of the model reasoning, making it difficult for the users to clearly distinguish among different types of model reasoning used.}
    \label{fig:appdx_id_test}
\end{figure}

Figure~\ref{fig:appdx_id_test} shows samples that illustrate the difficulty of clearly distinguishing model reasoning based on the feature attributions (i.e. identifiability). Along with Integrated Gradients (as in Figure~\ref{fig:identify}), all other methods including the ones presented here (Gradients, LRP-Z, DeepLIFT) highlight all objects in the image up to a certain level regardless of the model reasoning used, making it difficult to discern the reasoning based on the feature attributions.  

\newpage

\subsection{Results on Other Model Architectures}
\label{sec:appdx_other_architectures}

\begin{figure}[h]
    \centering
    \includegraphics[width=0.49\columnwidth]{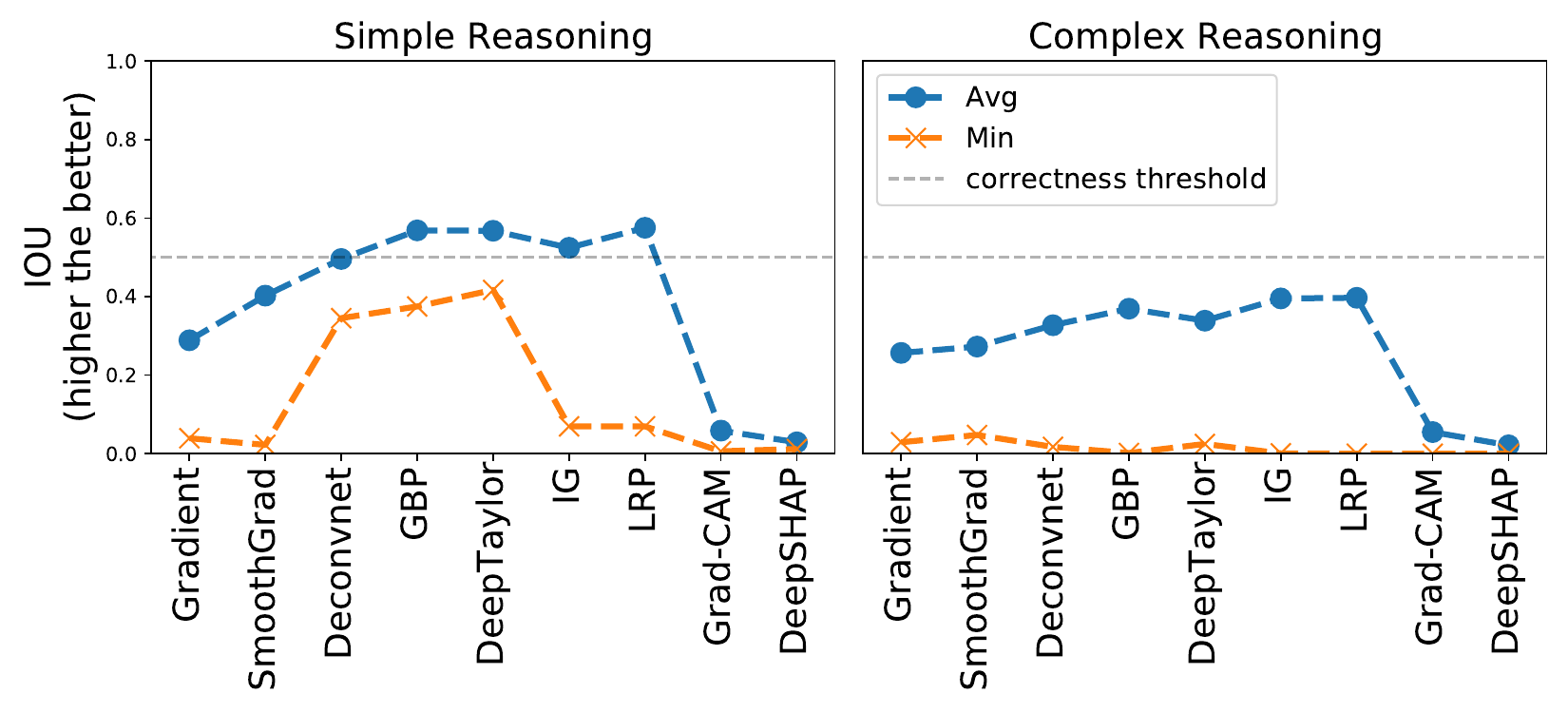}
    \includegraphics[width=0.49\columnwidth]{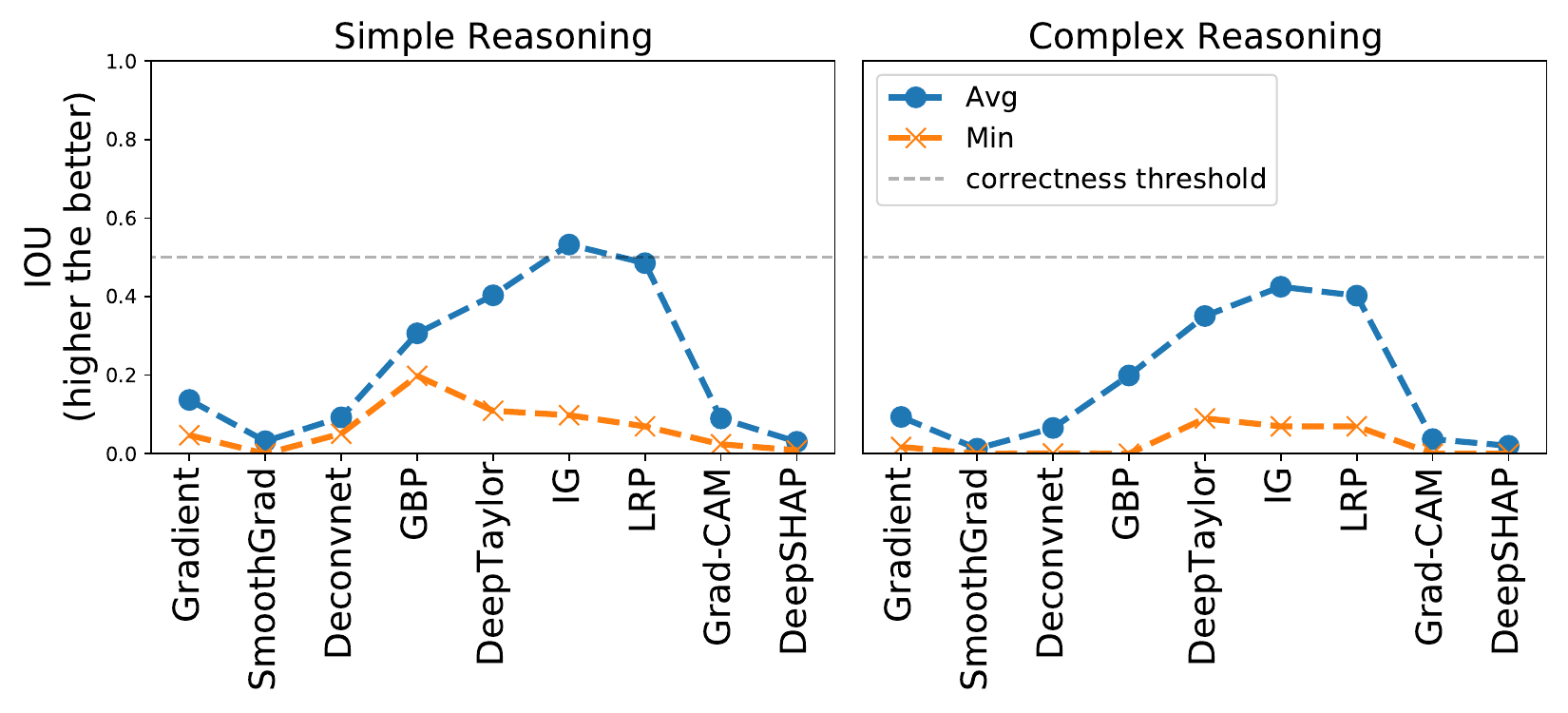}
    \caption{IOU results using AlexNet (left) and VGG16 (right) architecture for the model.}
    \label{fig:architecture_iou}
\end{figure}

\begin{figure}[h]
    \centering
    \includegraphics[width=0.49\textwidth]{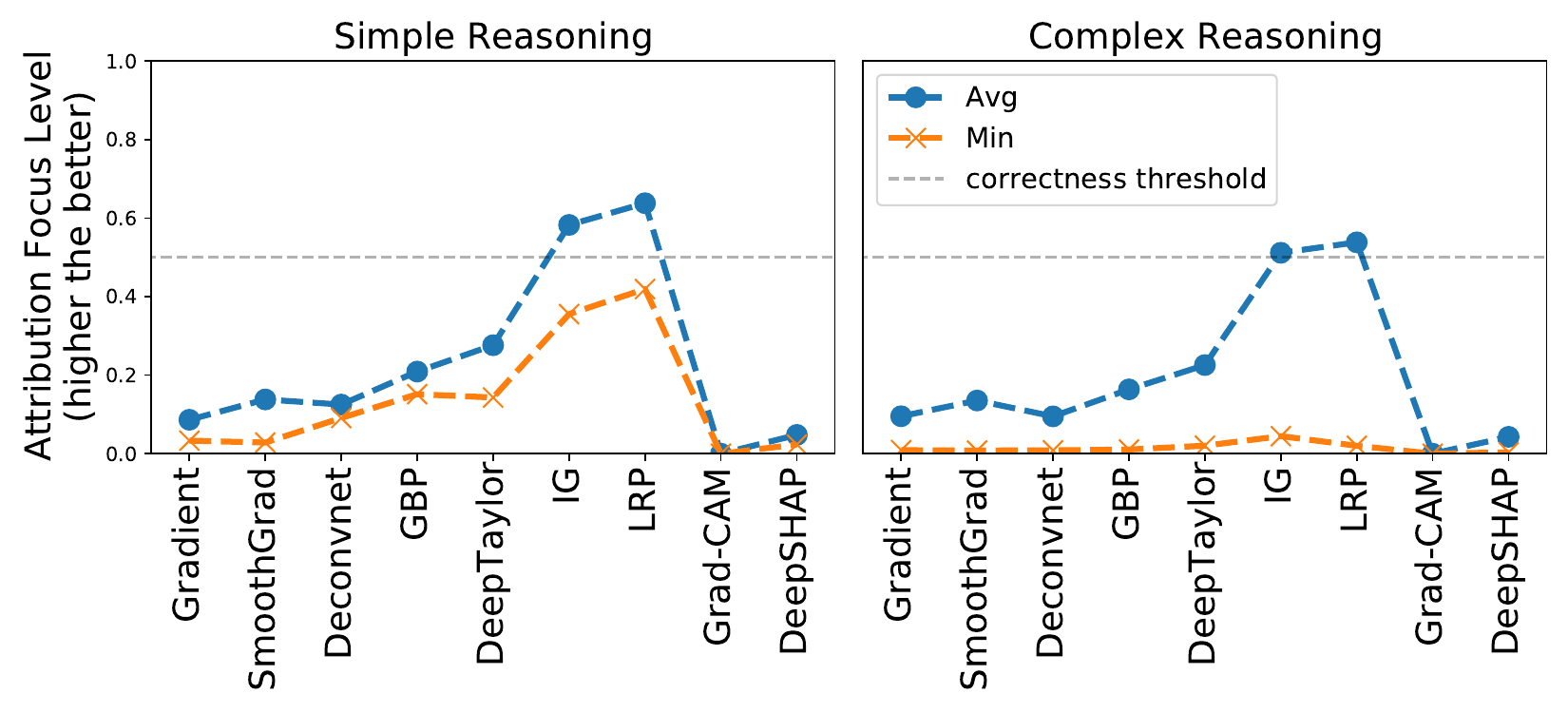}
    \includegraphics[width=0.49\textwidth]{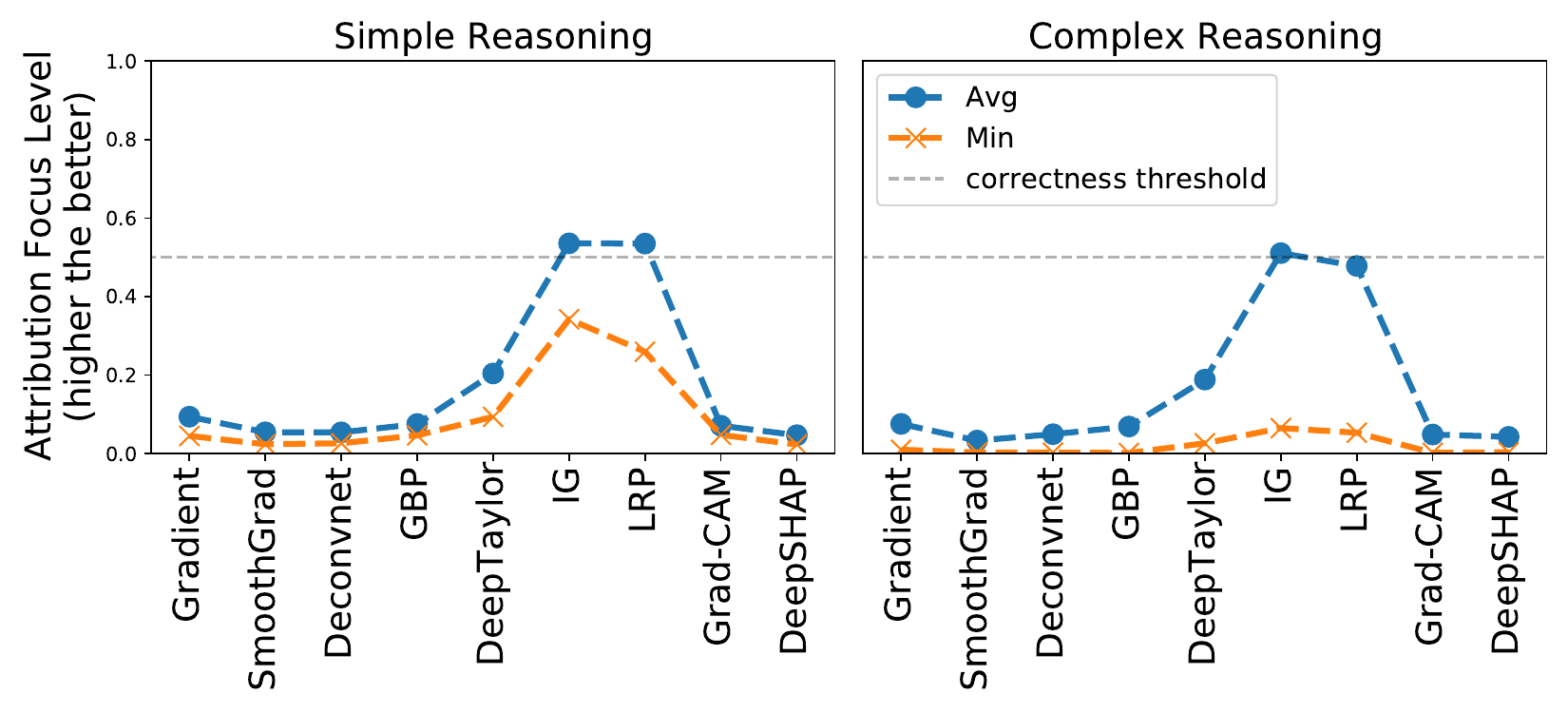}
    \caption{PAFL results using AlexNet (left) and VGG16 (right) architecture for the model.}
    \label{fig:architecture_afl}
\end{figure}

We repeat the same set of experiments with the model architecture using AlexNet~\cite{alexnet2012} and VGG16~\cite{simonyan2014very}, to confirm that the trends we observe for the saliency methods are not the artifact of architecture choice. IOU and AFL results for these models are shown in  Figures~\ref{fig:architecture_iou} and \ref{fig:architecture_afl}\footnote{Note that DeepLIFT was left out from the experiments on these models as the library for DeepLIFT does not support MaxPooling2D layer at the moment.}, which show similar trends we have observed so far in the simple CNN case: the average and the worst-case performance drops as the reasoning gets more complex.

\newpage

\subsection{Relationship with Adversarial Robustness}
\label{sec:appdx_adversarial}

\citet{shah2021input} showed that gradient feature attributions applied on more adversarially robust models tend to do better in ignoring the signals from spurious objects in the image. 
We verify that such trend for gradient attribution is somewhat true, yet the general problem persists for most of the other methods even for the robust models, in both simple and complex settings. 

\begin{figure}[h]
    \centering
    \includegraphics[width=0.49\textwidth]{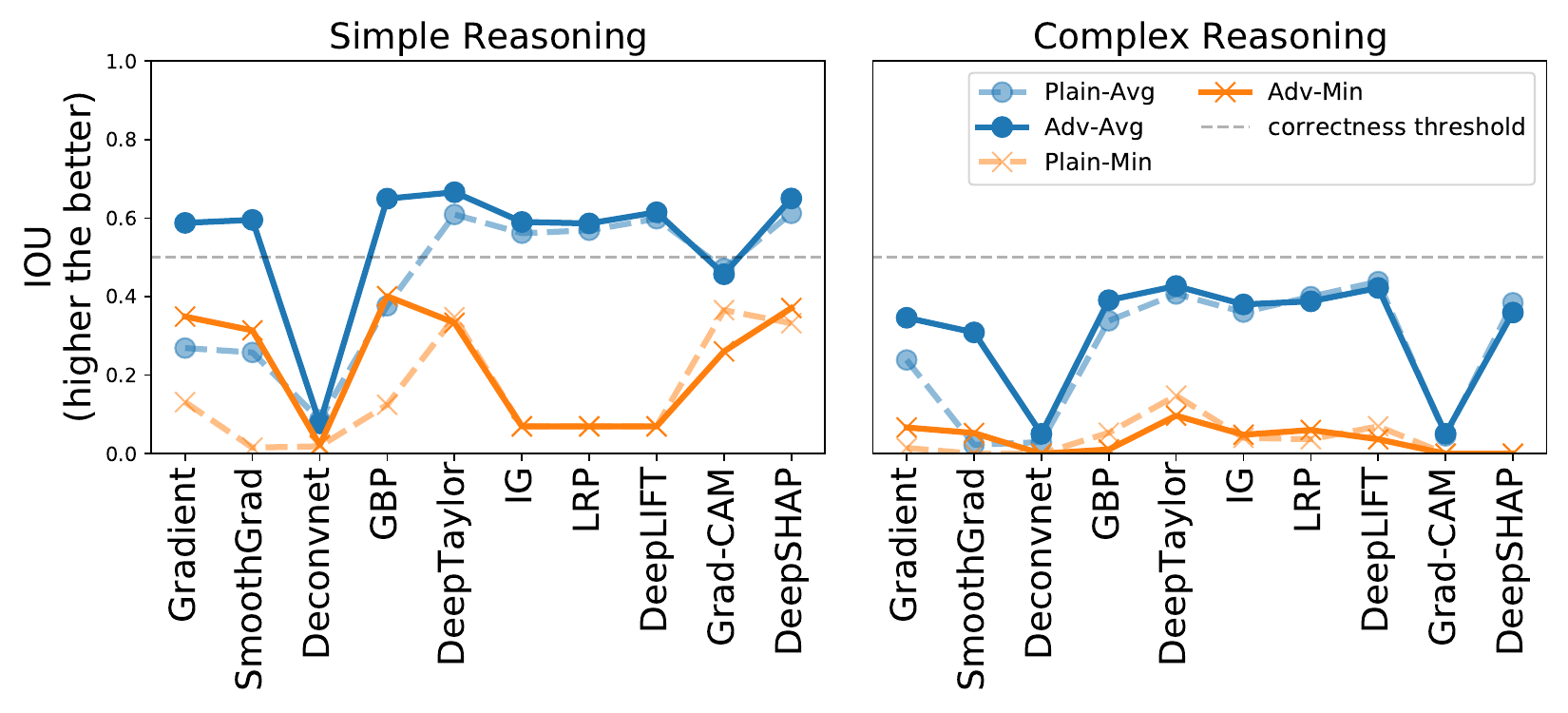}
    \includegraphics[width=0.49\textwidth]{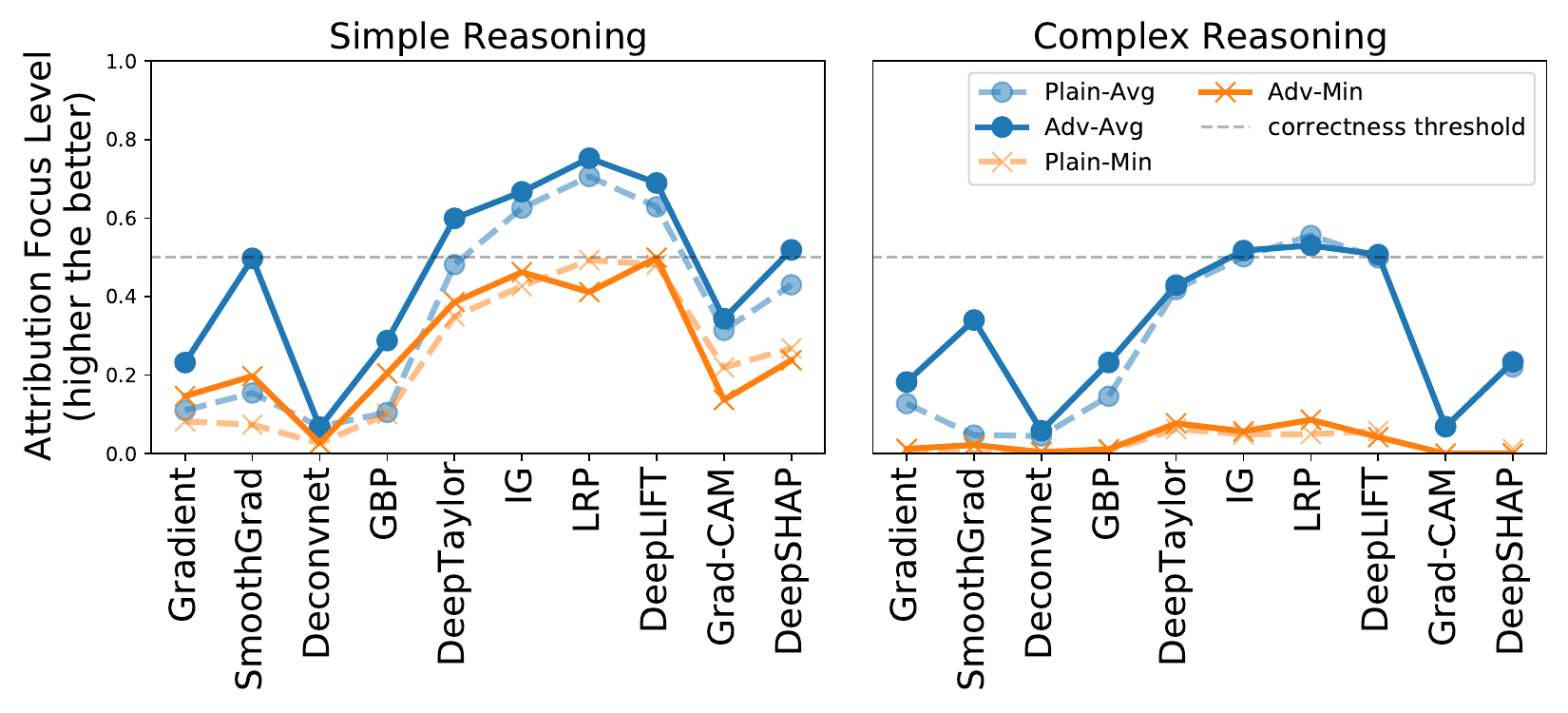}
    \caption{IOU (left) and PAFL (right) results on models trained against PGD attack (solid lines) and plain models (dotted lines, taken from Figure~\ref{fig:general_results}) for simple and complex reasoning settings. We observe increase in average performance for Gradient and SmoothGradient for simple reasoning compared to plain models, as suggested in \cite{shah2021input}. Nevertheless, in both metrics and across all methods, there still are sharp performance drops for complex reasoning.}
    \label{fig:adversarial_tests_all}
\end{figure}

Figure~\ref{fig:adversarial_tests_all} shows results for models trained against PGD attacks~\cite{madry2018towards}\footnote{Used a Python library https://github.com/Trusted-AI/adversarial-robustness-toolbox}. Throughout all methods, we still observe sharp performance drops in complex reasoning. 
We also observe the trend reported in \cite{shah2021input} as well, where Gradient and SmoothGradient's performance for simple reasoning (solid lines) is generally higher compared to plain models (dotted lines). 
As shown, while training the model against adversarial attacks can help some methods in simple reasoning settings, \NAME{} suggests that the overarching problem of methods not being able to reliably recover complex reasoning still persists.

\newpage

\subsection{\texttt{TextBox} with Noisy Background}
\label{sec:appdx_random_bg}

Instead of having zero-valued black pixels for the background, we set the background to consist of random pixel values between 0-150 for each of the RGB channels (before being normalized to [0,1]).

\begin{figure}[h]
    \centering
    \includegraphics[width=0.7\textwidth]{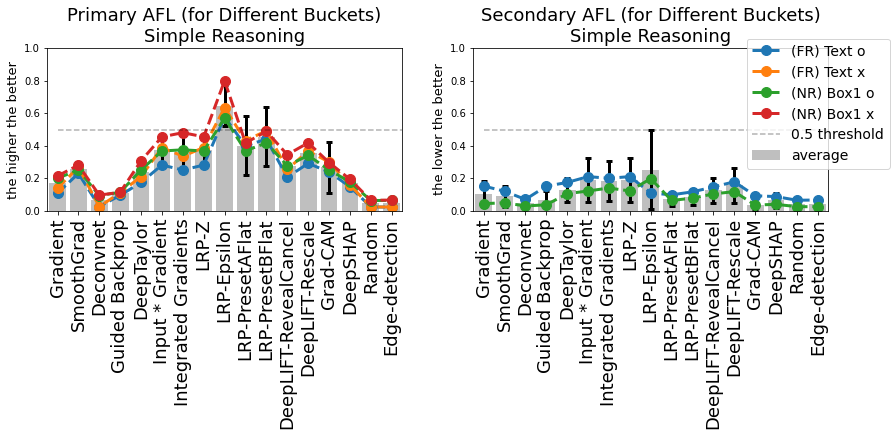}
    \caption{PAFL and SAFL for Simple Reasoning, tested with non-zero random noisy background.}
    \label{fig:simple-noisy}
\end{figure}

\begin{figure}[h]
    \begin{subfigure}[b]{0.72\textwidth}
         \centering
         \includegraphics[width=\textwidth]{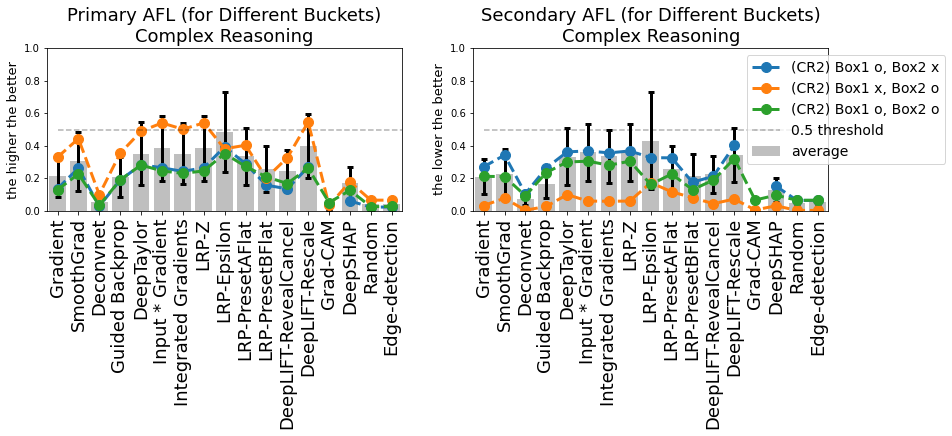}
         \caption{PAFL and SAFL for Complex Reasoning, tested with non-zero random background.}
         \label{fig:complex-noisy-bg}
     \end{subfigure} 
    \hfill
    \begin{subfigure}[b]{0.27\textwidth}
         \centering
         \includegraphics[width=\textwidth]{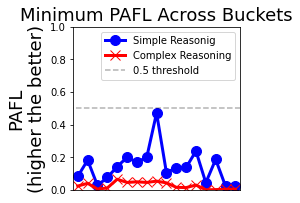}
         \caption{Minimum PAFL across buckets on simple (blue circle) vs. complex (red x) reasoning (x-axis is the same as Figure~\ref{fig:complex-noisy-bg}).}
         \label{fig:complex-noisy-pafl}
     \end{subfigure} 
    \caption{(a) PAFL and SAFL for complex reasoning. 
    (b) Worst-case buckets show worse PAFL values in complex reasoning. 
    }
    \label{fig:complex-noisy}
\end{figure}

The models were trained on these images to achieve near-perfect accuracy for all seven types of reasoning.  
Figure~\ref{fig:simple-noisy} shows PAFL and SAFL for simple reasoning settings, and Figure~\ref{fig:complex-noisy} for complex reasoning settings.
Similar to our earlier observations, methods under complex reasoning settings show sharp degradation of performance compared to the simple reasoning settings (lower PAFL, higher SAFL). 
Also we notice that additional noise from the background lowers the worst-case PAFL values throughout, even for the simple reasoning settings (Figure~\ref{fig:complex-noisy-pafl}). 

\newpage

\subsection{\texttt{TextBox} with Realistic Backgrounds}
\label{sec:appdx_changing_bg}

\begin{figure}[h]
    \centering
    \includegraphics[width=0.8\textwidth]{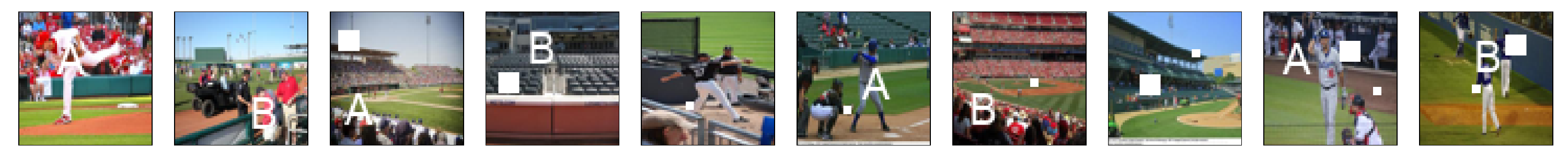}
    \caption{Images with real backgrounds used for the experiments.}
    \label{fig:real-background-examples}
\end{figure}

We replace the background of \texttt{TextBox} images with real images of different scenes taken from the Places dataset\footnote{\href{http://places2.csail.mit.edu/}{http://places2.csail.mit.edu/}}\cite{zhou2017places}. 
In particular, we replaced the background with images of baseball stadium and ran the same set of experiments (Figure~\ref{fig:real-background-examples}). The models trained on these images achieved near-perfect accuracy on both simple and complex settings (test accuracy for each reasoning: Simple-FR: 0.99, Simple-NR: 0.99, Complex-FR: 0.93, Complex-CR1: 0.98, Complex-CR2: 0.99, Complex-CR3: 0.93, Comlpex-CR4: 0.99).

\begin{figure}[h]
    \centering
    \includegraphics[width=0.6\textwidth]{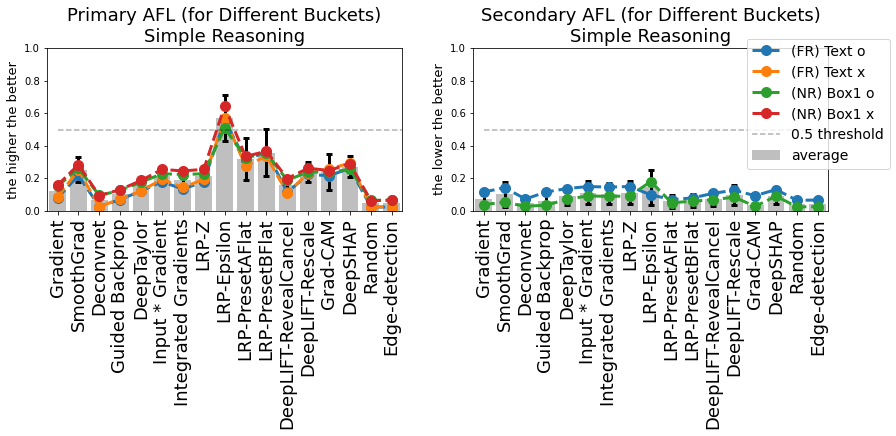}
    \caption{PAFL and SAFL for Simple Reasoning, for images with real baseball stadium as background.}
    \label{fig:bb-simple-afl}
\end{figure}

\begin{figure}[h]
    \begin{subfigure}[b]{0.7\textwidth}
         \centering
         \includegraphics[width=\textwidth]{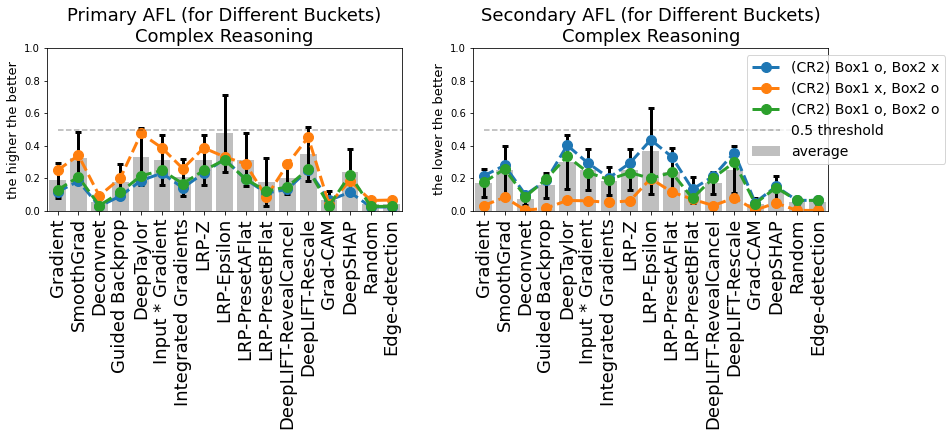}
         \caption{PAFL and SAFL for Complex Reasoning, tested with real background of baseball stadiums.}
         \label{fig:bb-complex-afl-values}
     \end{subfigure} 
    \hfill
    \begin{subfigure}[b]{0.25\textwidth}
         \centering
         \includegraphics[width=\textwidth]{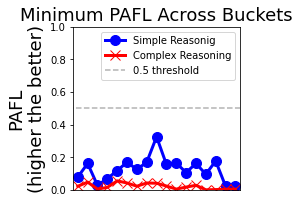}
         \caption{Minimum PAFL across buckets on simple (blue circle) vs. complex (red x) reasoning (x-axis is the same as Figure~\ref{fig:bb-complex-afl-values}).}
         \label{fig:bb-comparison}
     \end{subfigure} 
    \caption{(a) PAFL and SAFL for complex reasoning, for images with real baseball stadium as background.
    (b) Worst-case buckets show overall lower PAFL values in complex reasoning compared to simple reasoning.
    }
    \label{fig:bb-complex-afl}
\end{figure} 

Figure~\ref{fig:bb-simple-afl} and Figure~\ref{fig:bb-complex-afl} respectively shows AFL results on simple and complex reasoning settings. 
Figure~\ref{fig:bb-comparison} in particular shows that similar to the black background scenario studied earlier, there is a performance degradation moving from simple to complex reasoning, but at a lower level compared to the black background scenario.
Notably, the general performance drop relative to the black background setting was larger in this case than what we observed for the noisy background setting in Appendix~\ref{sec:appdx_random_bg}.
The results suggest that synthetic results of \NAME{} on the black background provides an optimistic upper bound for the methods' performance on the real background.


\end{document}